\pdfoutput=1

\documentclass[11pt]{article}

\usepackage[final]{acl}

\usepackage{times}
\usepackage{latexsym}

\usepackage[T1]{fontenc}

\usepackage[utf8]{inputenc}

\usepackage{microtype}

\usepackage{inconsolata}

\usepackage{graphicx}

\usepackage{microtype}
\usepackage{inconsolata}
\usepackage{hyperref}
\usepackage{booktabs}
\usepackage{adjustbox}
\usepackage{graphicx}
\usepackage{wrapfig}
\usepackage{amssymb}
\usepackage{color,xcolor,colortbl}
\usepackage{enumitem}
\usepackage{soul}
\usepackage{amsmath}
\usepackage{url}
\usepackage{algorithm}
\usepackage[noend]{algpseudocode}
\usepackage{booktabs}
\usepackage{bm,bbm}
\usepackage{mathtools}
\usepackage{array}
\usepackage{multirow}
\usepackage{multicol}
\usepackage{subcaption}
\usepackage{pbox}
\usepackage{pifont}
\usepackage{arydshln}
\usepackage{dblfloatfix}
\usepackage{relsize}
\usepackage{setspace}
\usepackage{tcolorbox}
\usepackage{listings}
\usepackage{setspace}
\tcbuselibrary{skins, breakable}
\newcommand\numberthis{\addtocounter{equation}{1}\tag{\theequation}}

\newenvironment{fontlmtt}{\fontfamily{lmtt}\selectfont}{\par} 

\usepackage[T1]{fontenc} 
\usepackage[scaled=.8]{beramono}

\definecolor{title}{HTML}{064A6C}
\definecolor{boxback}{HTML}{EFEDE1}
\definecolor{burgundy}{HTML}{660033}
\definecolor{bg-gray-50}{RGB}{249, 250, 251} 
\definecolor{bg-blue-600}{RGB}{37, 99, 235}  
\definecolor{text-blue-700}{RGB}{29, 78, 216} 
\definecolor{coolblack}{rgb}{0.0, 0.18, 0.39}
\definecolor{darkspringgreen}{rgb}{0.09, 0.45, 0.27}
\definecolor{flatgray}{RGB}{236, 240, 241}   
\definecolor{darkteel}{HTML}{1f6f6f}
\definecolor{lightteel}{HTML}{B3D9D9}
\definecolor{dark}{HTML}{064a6c}
\definecolor{light}{HTML}{efede1}

\newtcolorbox{promptbox}[2][]{
  floatplacement={#2},
  colframe=dark,colback=light!30!white,
  fonttitle=\small\ttfamily,
  fontupper=\small\ttfamily,
  title=#2,
  boxrule=0.5mm, 
  halign=flush left,
}

\definecolor{dark}{HTML}{064a6c}
\definecolor{light}{HTML}{efede1}

\title{\textsc{DeFine}: Decision-Making with Analogical Reasoning over Factor Profiles}

\author{Yebowen Hu,$^\dagger$ Xiaoyang Wang,$^\ddagger$ Wenlin Yao,$^\ddagger$ Yiming Lu,$^\S$ Daoan Zhang$^\star$\\ 
\textbf{Hassan Foroosh,$^\dagger$ Dong Yu,$^\ddagger$ Fei Liu$^\S$}\\[0.8em]
$^\dagger$University of Central Florida \, 
$^\ddagger$Tencent AI Lab, Seattle \\
$^\star$University of Rochester \,
$^\S$Emory University \\
\texttt{\{yebowen.hu, hassan.foroosh\}@ucf.edu} \;\; \texttt{daoan.zhang@rochester.edu}\\
\texttt{\{shawnxywang, wenlinyao, dyu\}@global.tencent.com \quad fei.liu@emory.edu}
}

\begin{document}
\maketitle
\begin{abstract}

LLMs are ideal for decision-making thanks to their ability to reason over long contexts. However, challenges arise when processing speech transcripts that describe complex scenarios, as they are verbose and include repetition, hedging, and vagueness. E.g., during a company's earnings call, an executive might project a positive revenue outlook to reassure investors, despite uncertainty regarding future earnings. It is crucial for LLMs to incorporate this uncertainty systematically when making decisions. In this paper, we introduce \textsc{DeFine}, a modular framework that constructs probabilistic factor profiles from complex scenarios. It then integrates these profiles with analogical reasoning, leveraging insights from similar past experiences to guide LLMs in making critical decisions in new situations. Our framework separates the tasks of quantifying uncertainty and incorporating it into LLM decision-making. This approach is particularly useful in areas such as consulting and financial deliberation, where making decisions under uncertainty is vital.

\end{abstract}

\section{Introduction}
\label{sec:intro}

\begin{figure*}
\centering
\includegraphics[width=6in]{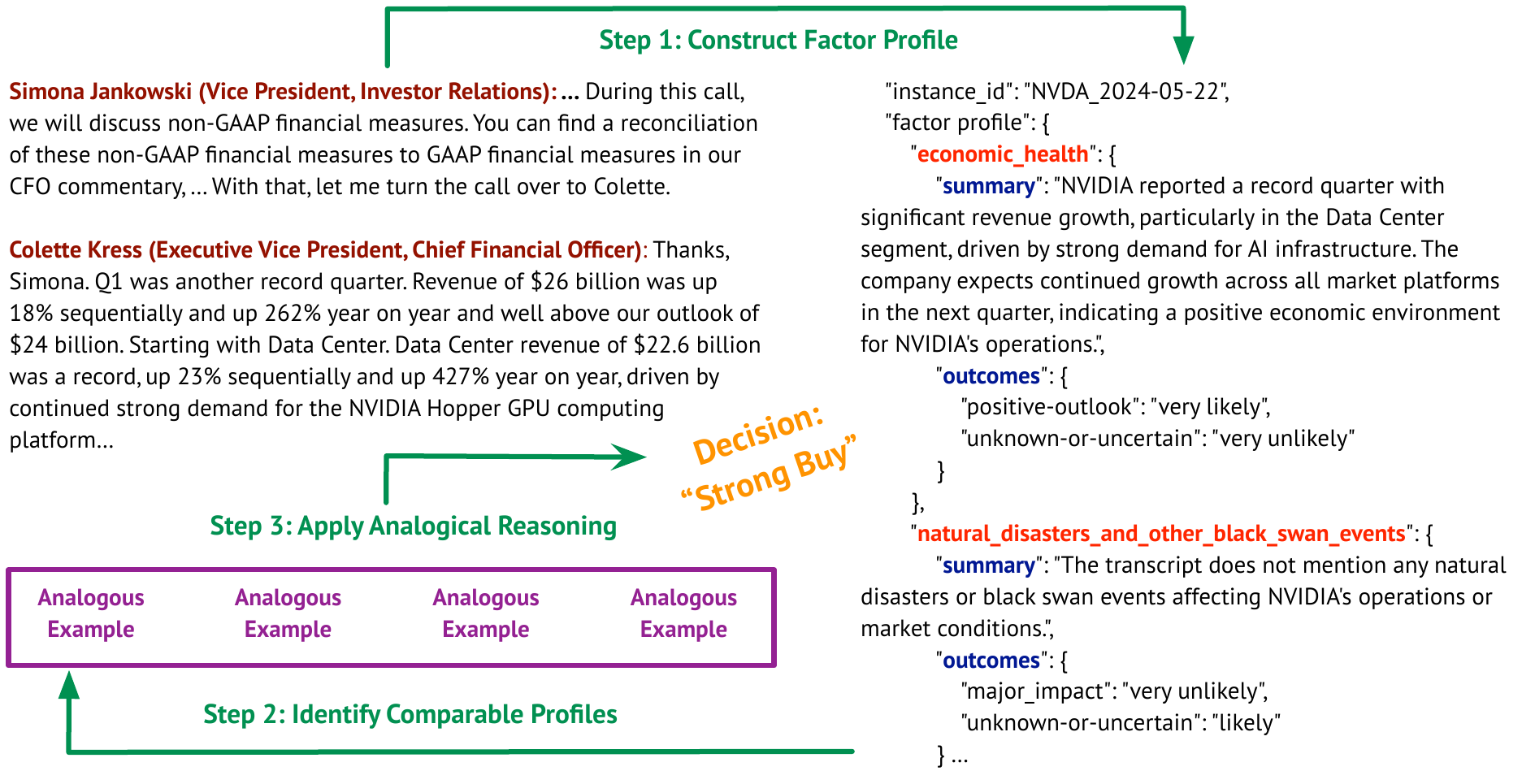}
\vspace{-0.05in}
\caption{An excerpt from a typical earnings call transcript and its associated factor profile.}
\label{fig:example}
\end{figure*}

Large language models are increasingly relied on for decision-making, thanks to their powerful reasoning capabilities~\cite{openai2024openaio1card,anthropic2025claude37,google2025gemini25}. 
While research has advanced rapidly in areas such as math, coding, and logical reasoning~\cite{bostrom-etal-2022-natural,huang2023reasoninglargelanguagemodels,sprague2024musr,mondorf2024accuracyevaluatingreasoningbehavior,li2025llmseasilylearnreason,ren2025deepseekproverv2advancingformalmathematical}, there is growing interest in exploring how LLMs reason through complex, real-world environments to make high-stakes decisions, such as financial investments~\citep{hsgac2024aihedgefunds}.

The challenges are compounded when processing long contexts~\citep{krishna-etal-2023-longeval,laban2024summaryhaystackchallengelongcontext}. Core issues may include recency bias, hallucinations, numerical inconsistencies, and more~\citep{liu2023lostmiddlelanguagemodels,hu-etal-2024-sportsmetrics,hu2024reasoningmeetsinformationaggregation,gao-etal-2024-insights}. While these models are designed to produce reasoning traces, their explanations can remain vague and even unfaithful~\cite{chen2025reasoningmodelsdontsay}. Critically, \emph{LLMs still lack precise, quantitative insight into the key factors that lead to their final decisions, as well as mechanisms for incorporating uncertainty into their decision-making.}

In this paper, we present \textsc{DeFine}, a new framework for constructing probabilistic factor profiles from speech transcripts that describe complex scenarios, and leveraging these profiles to enhance decision-making. For example, during an earnings call, a company executive might project strong revenue growth to boost investor confidence, despite substantial uncertainty surrounding these projections~\citep{mukherjee2022ectsumnewbenchmarkdataset}. As illustrated in Figure~\ref{fig:example}, \textsc{DeFine} generates a structured factor profile from each transcript, capturing not only what is explicitly stated, but also the implications of what is left unsaid. It then uses the BT model~\citep{10.2307/2334029} to identify dominant factors and evaluate how these factors collectively impact decision-making.

\textbf{\emph{Our research integrates probabilistic factor profiles with analogical reasoning}}, a type of reasoning that identifies connections between similar situations to facilitate knowledge transfer from a familiar context to a new situation~\citep{webb2023emergentanalogicalreasoninglarge,yasunaga2024largelanguagemodelsanalogical}. Instead of relying on text matching, we use factor profiles to retrieve analogous examples, which are historical cases with similar levels of uncertainty across key dimensions. Analogical reasoning further sets our work apart from traditional Bayesian inference models used in decision-making~\citep{halawi2024approachinghumanlevelforecastinglanguage,lin2024generatingconfidenceuncertaintyquantification,liu2024dellmaframeworkdecisionmaking}, which usually lack direct connections to historical cases. Our key contributions are as follows.

\begin{itemize}[topsep=0pt,itemsep=5pt,leftmargin=*]

\item We introduce \textsc{DeFine}, a modular framework that enhances LLM decision-making in complex scenarios. It transforms lengthy speech transcripts into structured factor profiles, which contain key decision factors and their associated uncertainties. Analogical reasoning retrieves comparable profiles and draws on similar past cases to make critical decisions. \textsc{DeFine} thereby adds a layer of transparency into the decision process.

\item Our research demonstrates how to effectively extract decision factors from lengthy transcripts and use them to forecast post-earnings stock movements. This approach has the potential to generalize to other domains, such as consulting, financial investments, and political debates~\citep{lehman-etal-2022-learning}, where discussions are complex and decisions carry significant consequences.

\end{itemize}

\section{The \textsc{DeFine} Framework}
\label{sec:define}

We investigate how LLMs make critical decisions about post-earnings stock movements using earnings call transcripts~\citep{Ni_2024,hsgac2024aihedgefunds}. An earnings call is a teleconference in which company executives discuss financial results with analysts and investors for a given quarter or fiscal year. As shown in Figure~\ref{fig:excerpt} in the Appendix, these transcripts typically consist of two parts: \emph{prepared remarks} from the company's executives and a subsequent \emph{Q\&A session}. During these calls, the executives provide a deep dive into the company's financials, discuss key performance indicators, and share strategic plans for the future. 

However, discussions of financials, e.g., revenue, expenses, and profit margins, can be overwhelming. A factor profile seeks to distills these discussions into multiple variables, allowing decision-makers to focus on the most impactful factors~\citep{eigner2024determinantsllmassisteddecisionmaking,feng2024birdtrustworthybayesianinference}. Notably, if critical elements such as debt levels are not addressed by company executives, they can be marked as ``unknown or uncertain.'' This approach contrasts with traditional textual summaries of the transcript~\citep{cho-etal-2021-streamhover,cho-etal-2022-toward,khatuya2024instructionguidedbulletpointsummarization}, which may be biased toward the topics emphasized by executives and discussed during the Q\&A session.

\subsection{Constructing Factor Profiles}
\label{sec:factor-profile}

Let $X$ be an earnings call transcript used to predict a stock investment decision $Y$ with five outcomes: strong buy, buy, hold, sell, and strong sell. We construct a \textbf{factor profile} for each transcript, defining a set of factors $\mathcal{F} = \{F_1, F_2, \dots, F_n\}$, where each factor $F_i$ has potential outcomes $\{O_{i1}, O_{i2}, \dots, O_{im}\}$.  
The likelihood of each outcome, given $X$, is modeled by a probabilistic function $P(O_{ij} | X)$. These probabilities are inferred using a methodology that optimally integrates textual reasoning with quantitative analysis. Thus, each factor outcome's probability informs the aggregation model that predicts $Y$. 

\begin{table}
\centering
\setlength{\tabcolsep}{2pt}
\renewcommand{\arraystretch}{0.9}
\begin{fontlmtt}
\begin{adjustbox}{max width=0.45\textwidth}
\begin{tabular}{ll}
\toprule
1. & Economic Health \\
2. & Market Sentiment and Investor Psychology \\
3. & Political Events and Government Policies \\
4. & Natural Disasters and Black Swan Events \\
5. & Geopolitical Issues \\
6. & Mergers and Major Acquisitions \\
7. & Regulatory Changes and Legal Issues \\
8. & Financial Health \\
9. & Company Growth \\
10. & Company Product Launches \\
11. & Supply Chain \\
12. & Technological Innovation \\
13. & Historical Earnings Per Share (EPS) \\
14. & Historical Revenue \\
15. & Historical Stock Prices \\
\bottomrule
\end{tabular}
\end{adjustbox}
\end{fontlmtt}
\vspace{-0.05in}
\caption{A curated set of 15 factors for forecasting stock movements following earnings.}
\label{tab:factors}
\end{table}

Our study focuses on 15 key factors grouped into three categories: \textit{macroeconomic influences} (e.g., economic health, market sentiment), \textit{company-specific dynamics} (e.g., mergers and major acquisitions, product launches), and \textit{historical financial metrics} (e.g., past earnings, stock prices). These factors were iteratively selected by querying the LLM for key variables in forecasting stock movements. We limit the set to 15 factors, each with two to three possible outcomes, ensuring a balance between complexity and performance while allowing future integration of analyst-identified factors. The full list of factors is in the Appendix.

We make use of the structured output capability of \texttt{GPT-4o-2024-08-06} to extract factor profiles from earnings call transcripts. Following the framework set by~\citet{liu2024dellmaframeworkdecisionmaking}, we provide the LLM with a list of factors, their potential outcomes, and associated verbalized likelihoods. For each factor, the analysis involves two steps: first, the LLM creates a concise summary specific to that factor from the transcript; second, it assigns a verbalized likelihood to each possible outcome, ranging from ``very unlikely'' to ``very likely.'' Specifically, the likelihoods of outcomes, such as EPS, revenue trends, and historical stock prices, are derived from the company's historical financial data. An example of the factor profile is shown in Figure~\ref{fig:example}, and the prompts used are detailed in the Appendix.

To convert these categorical likelihoods into probabilities, we employ the following normalization process: let $P_{i,j}$ denote the likelihood associated with the $j$-th outcome for the $i$-th factor. Here, verbalized likelihoods are converted to numerical values using the mapping \{very unlikely=1, unlikely=2, somewhat unlikely=3, somewhat likely=4, likely=5, very likely=6\}. Then, the probability $P(O_{ij} | X)$ is calculated as $P(O_{ij} | X) = \frac{P_{i,j}}{\sum_{k} P_{i,k}}$, ensuring the sum of outcomes for each factor equals 1. Alternative techniques, such as instructing the LLM to ``distribute 10 points among the outcomes'', have been explored~\citep{yang2024llmvotinghumanchoices}, our initial evaluation reveals that using verbalized likelihoods followed by normalization improves prediction accuracy compared to these direct probability distribution methods.

\subsection{Analyzing Key Factors Using the Bradley-Terry Model}
\label{sec:bradley-terry}

The Bradley-Terry model is a probabilistic framework used for estimating the relative strengths of items based on pairwise comparisons, and the outcome of each comparison indicates which of the two items is `better' in a specific context~\citep{10.2307/2334029}. This model has been widely used for ranking purposes in sports tournaments, LLM preference studies, and other domains where pairwise comparison data is available~\citep{hu2023decipherprefanalyzinginfluentialfactors,zhu2024principledreinforcementlearninghuman}. In this model, we estimate parameters that represent the strength of each factor. These parameters are generally presented on a logistic scale, where the probability that factor A is considered more significant than factor B is modeled as:
\begin{align*}
P(A > B) = \frac{e^{\beta_A}}{e^{\beta_A} + e^{\beta_B}}
\numberthis\label{eq:bradley-terry}
\end{align*}
Here, $\beta_A$ and $\beta_B$ represent the strengths of factors A and B, respectively. The estimated parameters are often exponentiated, so that $p_i = e^{\beta_i}$ measures the relative strength of each factor. A higher value indicates a stronger influence. In determining which factors to prioritize in a post-earnings analysis, those with higher Bradley-Terry scores are considered more crucial.

Consider a comparative analysis of two earnings call transcripts, A and B, transcript A is more likely to lead to favorable stock movements than transcript B (A $\succ$ B). We obtain such pairwise comparisons based on target labels; with `strong-buy' ranked higher than `hold', `sell', and `strong-sell'; `buy' outranking `sell' and `strong-sell'; and `hold' surpassing `strong-sell'. The comparison of A and B will involve creating a set of factor-outcome pairwise comparisons, where each outcome in transcript A is preferable to that in transcript B: $O_{\cdot,\cdot}^{(A)} \succ O_{\cdot,\cdot}^{(B)}$, suggesting that the factors associated with transcript A outperform those in transcript B.

We further consider the weight-adjusted effect of comparisons between factors. Our method compares the influence of factors from transcripts A and B by calculating an `expected occurrence', which is determined by multiplying the likelihood of these factors appearing in both transcripts, $P(O_{ij} | X^{(A)}) \times P(O_{ij} | X^{(B)})$. This approach provides a probability-based comparison, offering a more detailed evaluation than simple counting methods. These expected occurrences then feed into a Bradley-Terry model matrix $W$. The model helps to estimate the relative importance of each factor by assigning a coefficient $p_x$ to each outcome $O_{ij}$, indicating its influence on stock investment decisions. We refine these estimates using an EM-like algorithm, which iteratively adjusts and normalizes $p_x$ to best fit the observed data.
\begin{align*}
p'_x = W_x \left( \sum_{y \neq x} \frac{w_{xy} + w_{yx}}{p_x + p_y} \right)^{-1} 
\quad
p_x = \frac{p'_x}{\sum_{y=1}^M p'_y}
\numberthis\label{eq:bt_renorm}
\end{align*}

\section{Bayesian Decision-Making}
\label{sec:decision}
\vspace{-0.1in}
In Bayesian decision-making, utility functions play a crucial role in navigating uncertainty~\citep{halawi2024approachinghumanlevelforecastinglanguage,lin2024generatingconfidenceuncertaintyquantification,ye2024rationaldecisionmakingagentinternalized}. A Bayesian framework updates beliefs about possible outcomes. Decisions are then made by evaluating the expected utility for each possible action, which involves calculating the utility across the updated beliefs. This method ensures that choices are made to maximize expected utility, so decisions are aligned with the decision-maker's preferences and risk tolerance. 

Concretely, to compute $P(O_{ij} | X)$, we construct a probabilistic factor profile from a given earnings call transcript, where $O_{ij}$ represents the $j$-th outcome of the $i$-th factor. The likelihood $P(Y | O_{ij})$, which estimates how the $j$-th outcome influences stock investment decisions, is calculated using the Bradley-Terry model. This model provides a framework for quantifying the impact each factor outcome has on the decision-making process. Using these probabilities, the Bayesian decision-making formula integrates over all factors and their potential outcomes to determine the optimal action. The overall decision is derived by:
\begin{align*}
\hat{Y} = \arg\max_Y \sum_{i} \sum_{j} P(Y | O_{ij}) P(O_{ij} | X)
\numberthis\label{eq:bayesian}
\end{align*}
The parameters calculated by the Bradley-Terry model for $P(Y | O_{ij})$ help us determine how each factor influences stock movements. During our testing phase, transcripts are assigned to one of five decision categories based on their computed scores. For example, if the ground truth indicates there are $k$ `strong buy' recommendations, the top $k$ scoring transcripts are classified correspondingly as `strong buy'. This approach uses probabilistic factor profiles in conjunction with Bradley-Terry modeling to identify influential factors, providing a transparent method for understanding decision-driving elements. Moving forward, we extend beyond individual factors by examining analogous cases that directly influence decisions.

\section{Analogical Reasoning}
\label{sec:reasoning}

Analogical reasoning, which involves drawing parallels between similar situations~\citep{webb2023emergentanalogicalreasoninglarge,ozturkler2023thinksumprobabilisticreasoningsets,yuan2024analogykbunlockinganalogicalreasoning,sourati2024arnanalogicalreasoningnarratives,yasunaga2024largelanguagemodelsanalogical}, is an effective method for decision-making. This approach is particularly useful when analyzing how stocks react to earnings announcements by referencing past, similar events. For example, in the tech sector, stocks often show high volatility after earnings calls that introduce significant technological updates, even if the revenue and EPS meet expectations. If a tech company is rumored to discuss a new technology trend in its upcoming earnings announcement, using this method, we can infer that this company's stock might also experience increased volatility. Investors might use this analysis to make investment decisions or hedge against potential volatility.

Accurately identifying analogous examples from earnings call transcripts is crucial. We propose a method that utilizes probabilistic factor profiles, denoted as $P(O_{ij} | X)$, where $O_{ij}$ represents the $j$-th outcome of the $i$-th factor. To measure the similarity between profiles, we calculate the Kullback-Leibler (KL) divergence, which quantifies the information loss when one probability distribution approximates another. The KL divergence is computed as follows:
\begin{align*}
D_{KL}(P || Q) = \sum_{i=1}^n\sum_{j=1}^m P(O_{ij} | X) \log \frac{P(O_{ij} | X)}{Q(O_{ij} | X_c)}
\numberthis\label{eq:KLD}
\end{align*}
Here, $P$ represents the factor profile for the target transcript, and $Q$ denotes the profile for a comparative transcript $X_c$ from our training set. Transcripts with lower KL divergence values are considered more analogous, and therefore more likely to influence investor decisions similarly. 

During testing, we identify the Top-K profiles that show the least divergence from a test instance's profile and present these as analogical examples for the LLM to consider when reasoning about stock movements. The LLM is asked to select the most analogous example from the Top-K and carefully evaluates the current test instance to make its prediction. This approach ensures that the alignment between profiles is contextually appropriate, thereby drawing meaningful comparisons across different transcripts. By focusing on factor profiles rather than full transcripts or their summaries, we emphasize key market-moving information, avoiding unnecessary details. For example, Google and Broadcom could have analogous profiles even though their discussions in earnings calls might vary widely. Using factor profiles as analogous examples also requires significantly fewer tokens within the context window than full transcripts would.

\begin{table}
\centering
\setlength{\tabcolsep}{3pt}
\renewcommand{\arraystretch}{0.95}
\begin{adjustbox}{max width=0.44\textwidth}
\begin{tabular}{lr}
\textbf{Data Statistics} & \\
\toprule
Num. of Transcripts & 11,950 \\
Num. of Companies & 869 \\
Avg. \#Tokens per Transcript & 10,187 \\
Avg. \#QA Pairs per Transcript & 10 \\
Avg. \#Trans per Company & 14 \\
Avg. \#Speakers per Transcript & 12 \\
Year Range & 2017--2024\\
\bottomrule
\end{tabular}
\end{adjustbox}
\vspace{-0.05in}
\caption{Our dataset includes 11,950 earnings call transcripts from 800+ companies.}
\label{tab:data-stats}
\end{table}

\begin{table*}[t]
\setlength{\tabcolsep}{4pt}
\renewcommand{\arraystretch}{0.95}
\centering
\begin{tabular}{lrrrrllrrr}
\textbf{System} & \multicolumn{1}{c}{\textbf{Recall}} & \multicolumn{1}{c}{\textbf{Prec.}} & \multicolumn{1}{c}{\textbf{F$_1$}} & \multicolumn{1}{c}{\textbf{Accu.}} & & \textbf{Label} & \multicolumn{1}{c}{\textbf{Recall}} & \multicolumn{1}{c}{\textbf{Prec.}} & \multicolumn{1}{c}{\textbf{F$_1$}} \\
\cmidrule[0.8pt]{1-5}\cmidrule[0.8pt]{7-10}
LLM+CoT+Trans & 21.56 & 33.66 & 13.52 & 19.59 & & Strong Sell & 7.32 & 37.50 & 12.24\\
LLM+CoT+Summ & 22.77 & 16.17 & 14.12 & 20.61 & & Sell & 5.56 & 9.09 & 6.90 \\
LLM+CoT+Factors & 24.38 & 28.58 & 17.26 & 22.32 & & Hold & 29.84 & 28.24 & 29.02\\
DeLLMa {\footnotesize} & 38.30 & 23.14 & 16.68 & 22.35 & & Buy &  44.83 & 18.93 & 26.62 \\
\textbf{\textsc{DeFine} (Ours)} & 26.15 & 27.67 & \textbf{23.73} & \textbf{29.64} & & Strong Buy & 43.22 & 44.56 & \textbf{43.88}\\
\cmidrule[0.8pt]{1-5}\cmidrule[0.8pt]{7-10}
\end{tabular}
\vspace{-0.05in}
\caption{(\emph{Left}) We show the accuracy and macro-averaged F-scores for various systems. Our system, \textsc{DeFine}, which combines factor profiles with analogical reasoning, achieves the best performance. (\emph{Right}) \textsc{DeFine}'s performance across five categories: Strong Sell, Sell, Hold, Buy, and Strong Buy. }
\label{tab:results-GPT4o-30day}
\end{table*}

\begin{figure*}
\centering
\includegraphics[width=5.8in]{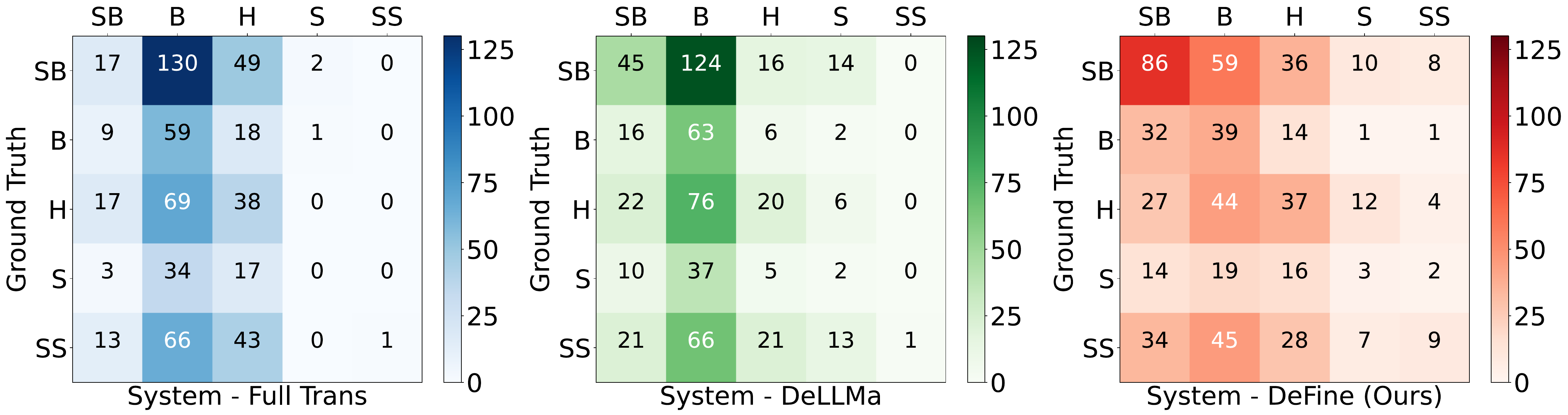}
\caption{A comparison of confusion matrices from the LLM+CoT+Trans, DeLLMa, and \textsc{DeFine} methods. While LLM+CoT+Trans and DeLLMa lean towards `Buy (B),' \textsc{DeFine} offers more balanced outcomes across all decision categories, showing notable improvement in `Strong Buy (SB),' `Buy (B),' `Hold (H),' and `Sell (S)' decisions.}
\label{fig:confusion_matrix}
\end{figure*}

\section{Data Collection}
\label{sec:dataset}

Our dataset contains 11,950 earnings call transcripts from S\&P 500 and NASDAQ 500 companies, gathered from the \href{fool.com}{Motley Fool} over the period of 2017--2024. The Motley Fool is a well-regarded financial service website that regularly publishes earnings call transcripts from U.S. companies. We make sure to follow their terms of use carefully during data collection. We do not use audio recordings or analyze acoustic or prosodic features. Each transcript is formatted as a JSON object, including the company's stock ticker, the date of the earnings announcement, participant names and their affiliations, executive prepared remarks, and a series of question-answer pairs from the Q\&A session.

Table~\ref{tab:data-stats} presents the statistics of our dataset. Each transcript averages 10,187 tokens and 133 sentences. They are sourced from 869 companies, each contributing an average of 14 transcripts. We obtain company stock prices from Yahoo Finance via the \texttt{yfinance} package and financial metrics such as revenue and earnings per share (EPS) from \href{alphaadvantage.co}{Alpha Advantage}. Our dataset spans from 2017 to 2024. It enhances previous studies which examined earnings call transcripts from 2002--2010~\citep{CIKM2020MAEC}; these earlier transcripts may already be used in LLM pretraining. To avoid data contamination, we established a new test set consisting of the most recent 587 transcripts from 2024, which are beyond the pretraining cut-off date for LLMs.

We seek to make stock investment decisions by analyzing earnings call transcripts and focusing on performance over the 30-day period. We establish the ground truth decision on the 30th day following each earnings announcement~\citep{sonkiya2021stockpricepredictionusing}: a stock drop exceeding 5\% corresponds to a `strong sell' decision, a decrease between 2\% and 5\% leads to a `sell', fluctuation within -2\% to +2\% is labeled `hold', an increase between 2\% and 5\% is labeled a `buy', and an increase above 5\% is a `strong buy'. In our test set, the distribution of these labels is as follows: `strong buy' at 34\%, `buy' at 15\%, `hold' at 21\%, `sell' at 9\%, and `strong sell' at 21\%. This distribution is generally balanced, reflecting a slightly bullish market trend in 2024.

\section{Experiments}
\label{sec:results}

In this section, we evaluate the decision-making performance of various systems, analyze the key factors that influence stock movement predictions, and conduct an analysis of analogical reasoning.

\begin{table*}[t]
\begin{minipage}[T]{0.47\textwidth}
\centering

\begin{adjustbox}{max width=\textwidth}
\begin{fontlmtt}
\setlength{\tabcolsep}{2pt}
\renewcommand{\arraystretch}{0.95}
\begin{tabular}{lc}
\textbf{Factor (Outcome)} & \textbf{Salience} \\
\toprule
- Regulatory changes and legal issues & 0.0364 \\
\quad happened (positive outlook) & \\
- Natural disasters and other black & 0.0360 \\
\quad swan events (major impact) & \\
- Political events and government & 0.0349 \\
\quad policies (major upheaval) & \\
- Geopolitical issues (escalation to & 0.0345 \\
\quad conflict)& \\
- Supply chain (positive outlook) & 0.0322 \\
- Tech innovation (positive outlook) & 0.0317 \\
- Historical stock price change (bullish) & 0.0316 \\
- Historical EPS (bullish) & 0.0315 \\
- Financial health (positive outlook) & 0.0311 \\
\bottomrule
\end{tabular}
\end{fontlmtt}
\end{adjustbox}
\vspace{-0.1in}
\caption{Influential factors that drive bullish investment decisions in the \textbf{\emph{Consumer Defensive}} sector, e.g., food and beverage, household products, and grocery stores.}
\label{tab:factors-consumer}

\end{minipage}
\hfill
\begin{minipage}[T]{0.5\textwidth}
\centering

\begin{adjustbox}{max width=1\textwidth}
\begin{fontlmtt}
\setlength{\tabcolsep}{1pt}
\renewcommand{\arraystretch}{0.95}
\begin{tabular}{lc}
\textbf{Factor (Outcome)} & \textbf{Salience} \\
\toprule
- Economic health (unknown or uncertain) & 0.0362 \\
- Market sentiment and investor & 0.0350  \\
\quad psychology (unknown or uncertain) &  \\
- Company growth (unknown or uncertain) & 0.0338 \\
- Supply chain (unknown or uncertain) & 0.0326 \\
- Geopolitical issues (escalation to conflict)  & 0.0322 \\
- Historical revenue (decline) & 0.0319 \\
- Historical stock price change (bullish) & 0.0318 \\
- Tech innovation (unknown or uncertain) & 0.0315 \\
- Natural disasters and other black & 0.0315 \\
\quad swan events (major impact) & \\
- Political events and government & 0.0313 \\
\quad policies (major upheaval) & \\
\bottomrule
\end{tabular}
\end{fontlmtt}
\end{adjustbox}
\vspace{-0.1in}
\caption{Factors that drive bullish investment decisions in the \textbf{\emph{Technology}} sector, including industry leaders such as Apple, Microsoft, Amazon, Google, and Meta.}
\label{tab:factors-technology}

\end{minipage}
\end{table*}

\begin{figure*}
\centering
\includegraphics[width=5in]{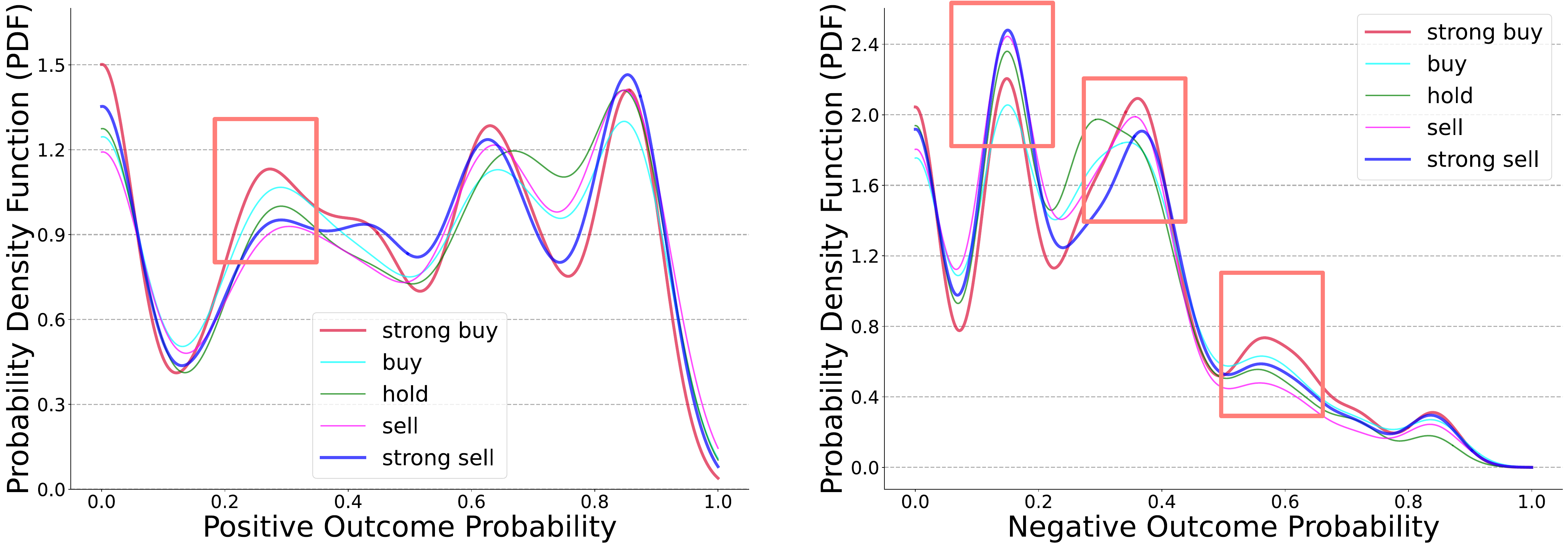}
\caption{We analyze and plot the probability density function (PDF) of positive and negative factor outcomes for different investment decisions. Highlighted sections illustrate where the gaps between strong buy (red) and strong sell (blue) decisions are most pronounced. }
\label{fig:outcome_likelihood}
\end{figure*}

\subsection{Decision-Making with DeFine}
\label{sec:system-comparisons}

We evaluate our system, \textsc{DeFine}, against various decision-making strategies: (a) \textbf{LLM+CoT+Trans}: The LLM processes the full earnings call transcript, using chain-of-thought reasoning to assign a label with interpretations. (b) \textbf{LLM+CoT+Summ} and \textbf{LLM+CoT+Factors}: Both follow a summarize-then-predict approach. \textbf{LLM+CoT+Summ} generates a textual summary, while \textbf{LLM+CoT+Factors} condenses information into a structured factor profile. Details on the prompts are in the Appendix.

Our system, \textsc{DeFine}, utilizes analogical reasoning by analyzing five analogous cases identified using KL-divergence as the distance metric. It examines these cases alongside the current factor profile to predict an appropriate label. In contrast, DeLLMa uses a decision theory approach and has shown strong performance in agriculture planning and finance~\citep{liu2024dellmaframeworkdecisionmaking}. For this approach, we pair each factor profile with possible labels and choose the top-ranked outcome as the final decision. 

In Table~\ref{tab:results-GPT4o-30day} (left), we present the accuracy and macro-averaged F-scores for various systems, all using \texttt{GPT-4o-2024-08-06}. Our new system, \textsc{DeFine}, which combines factor profiles with analogical reasoning, achieves the best performance. It surpasses the strong baseline system, DeLLMa, which involves ranking state-action pairs based on their preference levels as determined by the LLM. We find that LLMs generally make more accurate decisions when working with summaries rather than full transcripts; those transcripts typically contain around 10k tokens. This finding underscores the complexity of extracting and weighing key factors from lengthy transcripts, a task that remains challenging for most LLMs. In contrast, our factor profile method proves advantageous as it provides a balanced view of both macroeconomic factors and company-specific details, which are essential for rational decision-making. 

We further analyze \textsc{DeFine}'s performance across five categories: Strong Sell, Sell, Hold, Buy, and Strong Buy. Results are shown in Table~\ref{tab:results-GPT4o-30day} (right). \textsc{DeFine} performs best at `Strong Buy' recommendations and faces challenges with `Strong Sell' categories. This may be due to its reliance on earnings call transcripts, which often contain optimistic remarks from executives aimed at reassuring investors, potentially skewing predictions away from `Strong Sell.' Figure~\ref{fig:confusion_matrix} includes a comparison of confusion matrices from the LLM+CoT+Trans, DeLLMa, and \textsc{DeFine} methods. While LLM+CoT+Trans and DeLLMa predominantly lean towards `Buy,' \textsc{DeFine} offers more balanced outcomes across all decision categories, showing notable improvement in `Strong Buy,' `Buy,' `Hold,' and `Sell' decisions.

\subsection{Influential Factors}
\label{sec:key-factors}

We develop three variations of our \textsc{DeFine}-BT approach, each using the Bradley-Terry model for pairwise comparisons in different contexts: \textsc{DeFine}-BT-Same Sector compares companies within the same sector, \textsc{DeFine}-BT-Cross Sectors examines companies across different sectors, and \textsc{DeFine}-BT-Same Company analyzes a company's current earnings call transcript against its historical ones. To ensure fairness, we maintain the same number of pairwise comparisons across all three settings, downsampling where necessary. According to the F-scores presented in Table~\ref{tab:results-key-factors}, all \textsc{DeFine}-BT variants outperform both the random baseline, which assigns investment decisions randomly from five possible labels, and DeLLMa on the test set.

\begin{table*}
\setlength{\tabcolsep}{3.5pt}
\renewcommand{\arraystretch}{1.05}
\centering
\begin{adjustbox}{width=\textwidth}
\begin{fontlmtt}
\begin{small}
\begin{tabular}{lrrrrrrrrrrr}
 & \multicolumn{1}{c}{\textbf{Tech}} & \multicolumn{1}{c}{\textbf{FS}} & \multicolumn{1}{c}{\textbf{Health}} & \multicolumn{1}{c}{\textbf{CC}} & \multicolumn{1}{c}{\textbf{Ind}} & \multicolumn{1}{c}{\textbf{CS}} & \multicolumn{1}{c}{\textbf{CD}} & \multicolumn{1}{c}{\textbf{Energy}} & \multicolumn{1}{c}{\textbf{RE}} & \multicolumn{1}{c}{\textbf{BM}} & \multicolumn{1}{c}{\textbf{Util}} \\
\toprule
Technology (Tech) & \cellcolor{red!20}15.40 & 17.99 & 17.39 & 12.10 & 13.15 & 18.19 & \textbf{25.85} & \textbf{27.09} & 13.82 & 22.86 & 26.67 \\
Financial Services (FS) & 15.96 & \cellcolor{red!20}17.96 & \textbf{26.84} & 7.99 & 10.21 & 26.22 & 15.45 & 4.80 & 13.37 & 21.37 & 0.00 \\
Healthcare (Health) & 16.73 & 19.80 & \cellcolor{red!20}17.89 & 21.85 & 28.46 & 10.86 & 20.23 & 18.73 & 3.64 & \textbf{43.45} & \textbf{73.33} \\
Consumer Cyclical (CC) & 18.14 & 11.02 & 19.38 & \cellcolor{red!20}19.39 & 15.86 & 9.49 & 17.70 & 17.40 & 12.19 & 22.22 & 36.67 \\
Industrials (Ind) & 17.02 & 11.14 & 14.37 & 11.24 & \cellcolor{red!20}18.81 & 15.93 & 19.48 & 25.11 & 3.20 & 24.44 & 0.00 \\
Communication Services (CS) & 18.61 & 14.68 & 18.87 & 14.03 & 19.47 & \cellcolor{green!40}\textbf{33.70} & 10.71 & 16.99 & 11.87 & 10.26 & 13.33 \\
Consumer Defensive (CD) & \textbf{24.91} & 21.71 & 19.15 & 19.89 & 21.38 & 2.67 & \cellcolor{green!40}23.09 & 12.50 & 9.72 & 29.52 & 50.00 \\
Energy & 19.49 & 16.50 & 23.62 & 14.25 & 19.03 & 8.90 & 19.03 & \cellcolor{red!20}8.98 & 12.10 & 28.79 & 0.00 \\
Real Estate (RE) & 22.86 & 15.74 & 16.76 & 14.08 & 12.34 & 4.00 & 15.61 & 11.28 & \cellcolor{red!20}12.34 & 43.18 & 0.00 \\
Basic Materials (BM) & 20.67 & 13.69 & 15.26 & 18.18 & \textbf{29.64} & 9.52 & 21.19 & 17.19 & 17.10 & \cellcolor{red!20}16.67 & 37.50 \\
Utilities (Util) & 17.82 & \textbf{27.75} & 23.15 & \textbf{26.25} & 12.61 & 25.49 & 20.63 & 5.70 & \textbf{18.40} & 14.29 & \cellcolor{green!40}53.33 \\
\bottomrule
\end{tabular}
\end{small}
\end{fontlmtt}
\end{adjustbox}
\caption{The performance of \textsc{DeFine}-BT was evaluated by training it on one financial sector and testing it on another using 100 earnings call transcripts from each of the 11 sectors.
}
\label{tab:cross-sector}
\end{table*}

Among the three variants, \textsc{DeFine}-BT-Cross Sector achieves the highest scores in both F-Score and Accuracy. This indicates that considering pairwise comparisons between earnings announcements from a diverse range of companies can enhance predictions of stock movements. Table~\ref{tab:cross-sector} illustrates the performance of \textsc{DeFine}-BT-Cross Sector, which was trained on one sector and tested on another. For this analysis, 100 earnings call transcripts were selected from each of the 11 financial sectors: Technology, Healthcare, Financial Services, Consumer Defensive, Energy, Industrials, Utilities, Basic Materials, Real Estate, Consumer Cyclical, and Communication Services.

\begin{table}
    \centering
    \setlength{\tabcolsep}{2.8pt}
    \renewcommand{\arraystretch}{1}
    \begin{adjustbox}{max width=\textwidth}
    \begin{tabular}{lcc}
    & \textbf{F$_1$} & \textbf{Accu.} \\
    \toprule
    Random Baseline & 18.00 & 19.11 \\
    DeLLMa ({\footnotesize \citep{liu2024dellmaframeworkdecisionmaking}}) & 16.68 & 22.53 \\
    \textsc{DeFine}-BT-Same Sector & 20.11 & 22.15 \\
    \textsc{DeFine}-BT-Same Company & 20.42 & 23.68 \\
    \textsc{DeFine}-BT-Cross Sectors & \textbf{24.45} & \textbf{27.43} \\
    \bottomrule
    \end{tabular}
    \end{adjustbox}
    \caption{\textsc{DeFine}-BT-Cross Sector achieves the highest scores, suggesting that considering pairwise comparisons from a diverse range of companies can enhance the predictions of stock movements.}
    \label{tab:results-key-factors}
\end{table}

Tables~\ref{tab:factors-consumer} and~\ref{tab:factors-technology} highlight influential factors impacting investment decisions in the Consumer Defensive and Technology sectors, as identified by the Bradley-Terry model. In Consumer Defensive, which includes industries like food and beverage, household products, and grocery stores, significant drivers are natural disasters and black swan events, political events and government policies, and geopolitical issues. These challenging macroeconomic circumstances often lead to buy-in decisions from investors. In contrast, the Technology sector, with industry leaders such as Apple, Microsoft, Amazon, Google, Meta, and Nvidia, shows that decisions to invest often hinge on unclear or uncertain factors. Technology stocks have seen considerable growth from 2017--2024. This pattern suggests that investment models may favor purchases in these companies despite encountering negative issues in earnings announcements.

In Figure~\ref{fig:outcome_likelihood}, we analyze the probability of positive and negative factor outcomes, represented as a continuous random variable, and plot its probability density function (PDF) for various investment decisions. Highlighted sections illustrate where the gaps between strong buy (red) and strong sell (blue) decisions are most pronounced. Our analysis indicates that buy decisions often occur when the probability of positive outcomes is relatively low (about 0.2-0.3) and the likelihood of negative outcomes is moderate to high (ranging from 0.3 to 0.65), but not overly negative. Conversely, sell decisions tend to occur when negative outcome probabilities are minimal (about 0.1-0.2). These observations suggest that rational investment decisions can sometimes appear counterintuitive: essentially, selling high and buying low. We find that a thorough analysis of various factors is advantageous. Our approach incorporates not just the known issues but also the uncertain or hidden factors, thereby enhancing the decision-making process.

\subsection{Insights into Analogical Reasoning}
\label{sec:analogous-examples}

\begin{figure}
\centering
\includegraphics[width=2.5in]{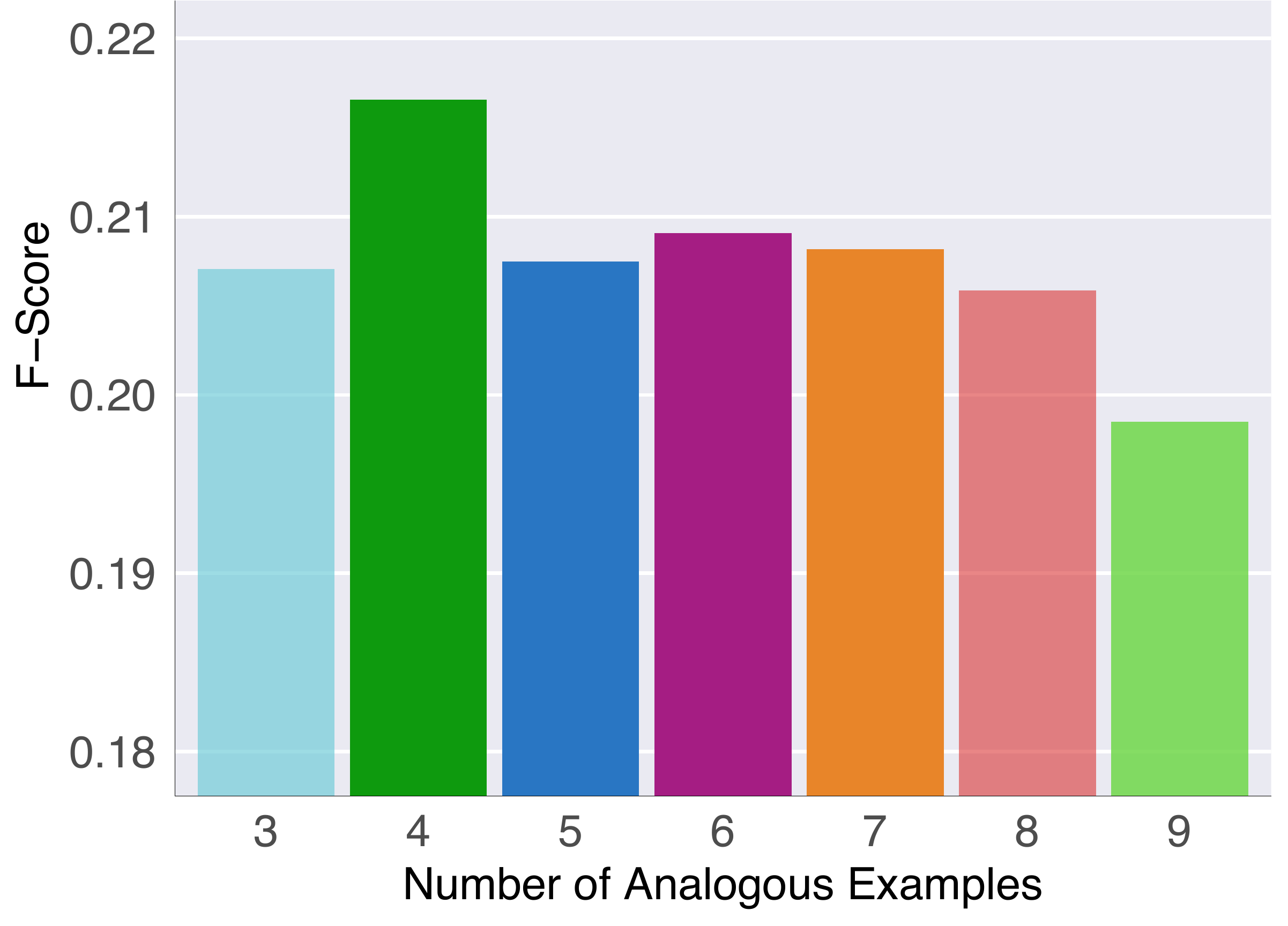}
\caption{Analogical reasoning in LLMs uses analogous examples for decision-making. We vary $K$ from 3 to 9 to assess its impact on the F-Score.
}
\label{fig:majority-vote}
\vspace{-0.15in}
\end{figure}

Analogical reasoning utilizes a select number of analogous examples, denoted as $K$, to inform decision-making in LLMs. In Figure~\ref{fig:majority-vote}, we adjust $K$ from 3 to 9 and observe its impact on the F-Score. In these experiments, we use the majority vote from the $K$ examples as the final prediction. We find that $K=4$ achieves the highest performance, potentially due to some tie-induced randomness compared to odd numbers. Typically, odd numbers for $K$ are preferred for majority voting to avoid ties, with $K=3, 5, 7$ showing similar effectiveness. For our system, \textsc{DeFine}, we have opted for $K=5$ to strike a balance between providing enough analogous examples and maintaining a manageable context length for the LLM.

Moreover, we examine how the most analogous examples influence \textsc{DeFine}'s predictions. Our study finds that in 69\% of cases, the LLM's predictions match the labels from the most analogous examples. In the other 31\% of cases, the LLM chooses to make its own predictions. E.g., when the analogous example is labeled ``Strong Buy,'' \textsc{DeFine} concurs with ``Strong Buy'' in 63\% of cases. It opts for ``Buy'' in 26\% and ``Hold'' in 11\% of the cases. Conversely, when the example is ``Strong Sell,'' \textsc{DeFine} agrees with ``Strong Sell'' 50\% of the time, chooses ``Sell'' in 25\% of cases, and ``Hold'' in 12.5\%. These results indicate that while \textsc{DeFine} effectively utilizes analogous historical data to inform its predictions, it also critically evaluates the current factor profiles, demonstrating a balanced approach in its decision-making abilities.

\section{Related Work}
\label{sec:related}

\paragraph{Analogical Reasoning}.\quad This type of reasoning identifies connections between similar, though not identical, situations to transfer knowledge from a known context to a new one~\citep{webb2023emergentanalogicalreasoninglarge,ozturkler2023thinksumprobabilisticreasoningsets,yu2024thoughtpropagationanalogicalapproach,yuan2024analogykbunlockinganalogicalreasoning,sourati2024arnanalogicalreasoningnarratives,yasunaga2024largelanguagemodelsanalogical}. It helps decision makers draw parallels between current situations and past experiences, effectively leveraging historical insights. Analogical reasoning plays a crucial role in various fields, e.g., doctors apply knowledge from one disease to diagnose another, and lawyers use past rulings to argue new cases~\citep{lehman-etal-2022-learning,charmet-etal-2022-complex,cao-etal-2024-pilot}. This ability to recognize and use similarities in different situations is important for decision-making.

While zero-shot analogical reasoning is a desired capability for LLMs, recent studies show they lack the robustness and generality of human analogy-making, as evidenced by counterexamples in tasks such as letter string analogies~\citep{hodel2024responseemergentanalogicalreasoning,lewis2024usingcounterfactualtasksevaluate}. \citet{musker2024semanticstructuremappingllmhuman} test both humans and LLMs on tasks that require transferring semantic structure and content between domains. \citet{yasunaga2024largelanguagemodelsanalogical} introduce analogical prompting, where LLMs self-generate relevant examples using prompts such as ``\emph{\# Recall relevant problems and solutions:}'' before solving the original problem; \citet{qin2024relevantrandomllmstruly} find that the accuracy of self-generated examples is key to eliciting such capability. Unlike previous research, our study employs probabilistic factor profiles to model analogical reasoning, grounding our approach in solid mathematical principles.

\paragraph{Decision-Making under Uncertainty.} The use of LLMs in decision-making has surged due to their remarkable ability to reason over complex scenarios~\citep{halawi2024approachinghumanlevelforecastinglanguage,lin2024generatingconfidenceuncertaintyquantification,ye2024rationaldecisionmakingagentinternalized,band2024linguisticcalibrationlongformgenerations}.
However, the challenge of balancing a multitude of often conflicting factors in decision making remains understudied. For example, \citet{falck2024incontextlearninglargelanguage} investigate whether adding more data points in in-context learning reduces uncertainty, as typically expected in Bayesian learning, and find evidence against this theory. The DeLLMa framework~\citep{liu2024dellmaframeworkdecisionmaking} incorporates uncertainty into LLM decision-making using Bayesian networks and has been tested on tasks such as agriculture planning and finance. \citet{feng2024birdtrustworthybayesianinference} employ LLM entailment to map factors to context and utilize trained Bayesian models for probability estimation. Our work builds on these initiatives by integrating analogical reasoning with factor profiles to enhance the accuracy and transparency of LLM decision-making.

\paragraph{Financial Forecasting.}\quad Recent advancements in LLMs have revolutionized traditional financial tasks~\citep{keith-stent-2019-modeling,sawhney-etal-2020-voltage,sawhney-etal-2021-multimodal,chuang-yang-2022-buy,ang-lim-2022-guided,sang-bao-2022-dialoguegat,medya2022exploratorystudystockprice,wang2023methodsacquiringincorporatingknowledge,Koa_2024,srivastava2024evaluatingllmsmathematicalreasoning}. Notably, \citet{chen2022finqadatasetnumericalreasoning} introduce FinQA, a dataset constructed from financial statements for assessing LLMs' multi-step numerical reasoning. Moreover, TAT-QA~\citep{zhu2021tatqa} tackles QA over tabular and textual data; FiNER~\citep{Loukas_2022} focuses on numerical entity recognition; DocFinQA~\citep{reddy2024docfinqa} is a dataset designed for long-document financial QA; RiskLabs~\citep{cao2024risklabspredictingfinancialrisk} employs LLMs for financial risk assessments. \citet{nie2024surveylargelanguagemodels} provide a comprehensive survey on the use of LLMs across various financial domains. Our study focuses on analyzing earnings transcripts to understand how LLMs handle the ambiguities inherent in spoken language, thus providing insight into their decision-making under uncertainty. The research findings have broader applications including medical consultations, negotiations, and political debates.

\section{Conclusion}
\label{sec:conclusion}

We propose \textsc{DeFine}, a new framework for decision-making in complex scenarios, such as those encountered in corporate earnings calls. By combining probabilistic factor profiles with analogical reasoning, this framework not only captures the uncertainties embedded in earnings call transcripts but also allows the LLM to apply previous insights to new challenges more efficiently. Our approach surpasses strong baseline models and enhances the practical utility of LLMs by identifying analogous examples. The \textsc{DeFine} framework offers a promising avenue for navigating complex data and supporting decision-making processes.

\section*{Acknowledgements}
We are grateful to the reviewers for their insightful feedback, which has helped improve our paper. This research has been partially supported by the NSF CAREER award, \#2303655.

\section{Limitations}
\label{sec:limitations}

The \textsc{DeFine} framework, as detailed in our paper, offers a promising approach to enhancing decision-making through its innovative use of probabilistic factor profiles and analogical reasoning. Developed under carefully controlled experimental conditions, its potential is noteworthy. However, it's important to acknowledge that its performance might vary in the complexity of real-world settings due to factors such as data quality and specific contextual nuances. We encourage users to consider these variables when adapting the \textsc{DeFine} approach to ensure its optimal application and to mitigate potential discrepancies between expected and actual results.

\bibliography{custom,anthology}

\begin{thebibliography}{66}
\providecommand{\natexlab}[1]{#1}

\bibitem[{Ang and Lim(2022)}]{ang-lim-2022-guided}
Gary Ang and Ee-Peng Lim. 2022.
\newblock \href {https://doi.org/10.18653/v1/2022.acl-long.437} {Guided attention multimodal multitask financial forecasting with inter-company relationships and global and local news}.
\newblock In \emph{Proceedings of the 60th Annual Meeting of the Association for Computational Linguistics (Volume 1: Long Papers)}, pages 6313--6326, Dublin, Ireland. Association for Computational Linguistics.

\bibitem[{Anthropic(2025)}]{anthropic2025claude37}
Anthropic. 2025.
\newblock \href {https://www.anthropic.com/news/claude-3-7-sonnet} {Claude 3.7 sonnet and claude code}.
\newblock Technical report.
\newblock Accessed: 2025-05-13.

\bibitem[{Band et~al.(2024)Band, Li, Ma, and Hashimoto}]{band2024linguisticcalibrationlongformgenerations}
Neil Band, Xuechen Li, Tengyu Ma, and Tatsunori Hashimoto. 2024.
\newblock \href {https://arxiv.org/abs/2404.00474} {Linguistic calibration of long-form generations}.
\newblock \emph{Preprint}, arXiv:2404.00474.

\bibitem[{Bostrom et~al.(2022)Bostrom, Sprague, Chaudhuri, and Durrett}]{bostrom-etal-2022-natural}
Kaj Bostrom, Zayne Sprague, Swarat Chaudhuri, and Greg Durrett. 2022.
\newblock \href {https://doi.org/10.18653/v1/2022.findings-emnlp.358} {Natural language deduction through search over statement compositions}.
\newblock In \emph{Findings of the Association for Computational Linguistics: EMNLP 2022}, pages 4871--4883, Abu Dhabi, United Arab Emirates. Association for Computational Linguistics.

\bibitem[{Bradley and Terry(1952)}]{10.2307/2334029}
Ralph~Allan Bradley and Milton~E. Terry. 1952.
\newblock \href {http://www.jstor.org/stable/2334029} {Rank analysis of incomplete block designs: I. the method of paired comparisons}.
\newblock \emph{Biometrika}, 39(3/4):324--345.

\bibitem[{Cao et~al.(2024{\natexlab{a}})Cao, Wang, Xiao, and Sun}]{cao-etal-2024-pilot}
Lang Cao, Zifeng Wang, Cao Xiao, and Jimeng Sun. 2024{\natexlab{a}}.
\newblock \href {https://doi.org/10.18653/v1/2024.naacl-long.34} {{PILOT}: Legal case outcome prediction with case law}.
\newblock In \emph{Proceedings of the 2024 Conference of the North American Chapter of the Association for Computational Linguistics: Human Language Technologies (Volume 1: Long Papers)}, pages 609--621, Mexico City, Mexico. Association for Computational Linguistics.

\bibitem[{Cao et~al.(2024{\natexlab{b}})Cao, Chen, Pei, Dimino, Ausiello, Kumar, Subbalakshmi, and Ndiaye}]{cao2024risklabspredictingfinancialrisk}
Yupeng Cao, Zhi Chen, Qingyun Pei, Fabrizio Dimino, Lorenzo Ausiello, Prashant Kumar, K.~P. Subbalakshmi, and Papa~Momar Ndiaye. 2024{\natexlab{b}}.
\newblock \href {https://arxiv.org/abs/2404.07452} {Risklabs: Predicting financial risk using large language model based on multi-sources data}.
\newblock \emph{Preprint}, arXiv:2404.07452.

\bibitem[{Charmet et~al.(2022)Charmet, Cherichi, Allain, Czerwinska, Fouret, Sagot, and Bawden}]{charmet-etal-2022-complex}
Thibault Charmet, In{\`e}s Cherichi, Matthieu Allain, Urszula Czerwinska, Amaury Fouret, Beno{\^\i}t Sagot, and Rachel Bawden. 2022.
\newblock \href {https://aclanthology.org/2022.lrec-1.509} {Complex labelling and similarity prediction in legal texts: Automatic analysis of {F}rance{'}s court of cassation rulings}.
\newblock In \emph{Proceedings of the Thirteenth Language Resources and Evaluation Conference}, pages 4754--4766, Marseille, France. European Language Resources Association.

\bibitem[{Chen et~al.(2025)Chen, Benton, Radhakrishnan, Uesato, Denison, Schulman, Somani, Hase, Wagner, Roger, Mikulik, Bowman, Leike, Kaplan, and Perez}]{chen2025reasoningmodelsdontsay}
Yanda Chen, Joe Benton, Ansh Radhakrishnan, Jonathan Uesato, Carson Denison, John Schulman, Arushi Somani, Peter Hase, Misha Wagner, Fabien Roger, Vlad Mikulik, Samuel~R. Bowman, Jan Leike, Jared Kaplan, and Ethan Perez. 2025.
\newblock \href {https://arxiv.org/abs/2505.05410} {Reasoning models don't always say what they think}.
\newblock \emph{Preprint}, arXiv:2505.05410.

\bibitem[{Chen et~al.(2022)Chen, Chen, Smiley, Shah, Borova, Langdon, Moussa, Beane, Huang, Routledge, and Wang}]{chen2022finqadatasetnumericalreasoning}
Zhiyu Chen, Wenhu Chen, Charese Smiley, Sameena Shah, Iana Borova, Dylan Langdon, Reema Moussa, Matt Beane, Ting-Hao Huang, Bryan Routledge, and William~Yang Wang. 2022.
\newblock \href {https://arxiv.org/abs/2109.00122} {Finqa: A dataset of numerical reasoning over financial data}.
\newblock \emph{Preprint}, arXiv:2109.00122.

\bibitem[{Cho et~al.(2021)Cho, Dernoncourt, Ganter, Bui, Lipka, Chang, Jin, Brandt, Foroosh, and Liu}]{cho-etal-2021-streamhover}
Sangwoo Cho, Franck Dernoncourt, Tim Ganter, Trung Bui, Nedim Lipka, Walter Chang, Hailin Jin, Jonathan Brandt, Hassan Foroosh, and Fei Liu. 2021.
\newblock \href {https://doi.org/10.18653/v1/2021.emnlp-main.520} {{S}tream{H}over: Livestream transcript summarization and annotation}.
\newblock In \emph{Proceedings of the 2021 Conference on Empirical Methods in Natural Language Processing}, pages 6457--6474, Online and Punta Cana, Dominican Republic. Association for Computational Linguistics.

\bibitem[{Cho et~al.(2022)Cho, Song, Wang, Liu, and Yu}]{cho-etal-2022-toward}
Sangwoo Cho, Kaiqiang Song, Xiaoyang Wang, Fei Liu, and Dong Yu. 2022.
\newblock \href {https://doi.org/10.18653/v1/2022.emnlp-main.8} {Toward unifying text segmentation and long document summarization}.
\newblock In \emph{Proceedings of the 2022 Conference on Empirical Methods in Natural Language Processing}, pages 106--118, Abu Dhabi, United Arab Emirates. Association for Computational Linguistics.

\bibitem[{Chuang and Yang(2022)}]{chuang-yang-2022-buy}
Chengyu Chuang and Yi~Yang. 2022.
\newblock \href {https://doi.org/10.18653/v1/2022.acl-short.12} {Buy tesla, sell ford: Assessing implicit stock market preference in pre-trained language models}.
\newblock In \emph{Proceedings of the 60th Annual Meeting of the Association for Computational Linguistics (Volume 2: Short Papers)}, pages 100--105, Dublin, Ireland. Association for Computational Linguistics.

\bibitem[{Eigner and Händler(2024)}]{eigner2024determinantsllmassisteddecisionmaking}
Eva Eigner and Thorsten Händler. 2024.
\newblock \href {https://arxiv.org/abs/2402.17385} {Determinants of llm-assisted decision-making}.
\newblock \emph{Preprint}, arXiv:2402.17385.

\bibitem[{Falck et~al.(2024)Falck, Wang, and Holmes}]{falck2024incontextlearninglargelanguage}
Fabian Falck, Ziyu Wang, and Chris Holmes. 2024.
\newblock \href {https://arxiv.org/abs/2406.00793} {Is in-context learning in large language models bayesian? a martingale perspective}.
\newblock \emph{Preprint}, arXiv:2406.00793.

\bibitem[{Feng et~al.(2024)Feng, Zhou, Lin, and Roth}]{feng2024birdtrustworthybayesianinference}
Yu~Feng, Ben Zhou, Weidong Lin, and Dan Roth. 2024.
\newblock \href {https://arxiv.org/abs/2404.12494} {Bird: A trustworthy bayesian inference framework for large language models}.
\newblock \emph{Preprint}, arXiv:2404.12494.

\bibitem[{Gao et~al.(2024)Gao, Lu, Yu, Byerly, and Khashabi}]{gao-etal-2024-insights}
Muhan Gao, TaiMing Lu, Kuai Yu, Adam Byerly, and Daniel Khashabi. 2024.
\newblock \href {https://doi.org/10.18653/v1/2024.findings-emnlp.447} {Insights into {LLM} long-context failures: When transformers know but don`t tell}.
\newblock In \emph{Findings of the Association for Computational Linguistics: EMNLP 2024}, pages 7611--7625, Miami, Florida, USA. Association for Computational Linguistics.

\bibitem[{Halawi et~al.(2024)Halawi, Zhang, Yueh-Han, and Steinhardt}]{halawi2024approachinghumanlevelforecastinglanguage}
Danny Halawi, Fred Zhang, Chen Yueh-Han, and Jacob Steinhardt. 2024.
\newblock \href {https://arxiv.org/abs/2402.18563} {Approaching human-level forecasting with language models}.
\newblock \emph{Preprint}, arXiv:2402.18563.

\bibitem[{Hodel and West(2024)}]{hodel2024responseemergentanalogicalreasoning}
Damian Hodel and Jevin West. 2024.
\newblock \href {https://arxiv.org/abs/2308.16118} {Response: Emergent analogical reasoning in large language models}.
\newblock \emph{Preprint}, arXiv:2308.16118.

\bibitem[{Hu et~al.(2023)Hu, Song, Cho, Wang, Foroosh, and Liu}]{hu2023decipherprefanalyzinginfluentialfactors}
Yebowen Hu, Kaiqiang Song, Sangwoo Cho, Xiaoyang Wang, Hassan Foroosh, and Fei Liu. 2023.
\newblock \href {https://arxiv.org/abs/2305.14702} {Decipherpref: Analyzing influential factors in human preference judgments via gpt-4}.
\newblock \emph{Preprint}, arXiv:2305.14702.

\bibitem[{Hu et~al.(2024{\natexlab{a}})Hu, Song, Cho, Wang, Foroosh, Yu, and Liu}]{hu-etal-2024-sportsmetrics}
Yebowen Hu, Kaiqiang Song, Sangwoo Cho, Xiaoyang Wang, Hassan Foroosh, Dong Yu, and Fei Liu. 2024{\natexlab{a}}.
\newblock \href {https://doi.org/10.18653/v1/2024.acl-long.17} {{S}ports{M}etrics: Blending text and numerical data to understand information fusion in {LLM}s}.
\newblock In \emph{Proceedings of the 62nd Annual Meeting of the Association for Computational Linguistics (Volume 1: Long Papers)}, pages 267--278, Bangkok, Thailand. Association for Computational Linguistics.

\bibitem[{Hu et~al.(2024{\natexlab{b}})Hu, Song, Cho, Wang, Yao, Foroosh, Yu, and Liu}]{hu2024reasoningmeetsinformationaggregation}
Yebowen Hu, Kaiqiang Song, Sangwoo Cho, Xiaoyang Wang, Wenlin Yao, Hassan Foroosh, Dong Yu, and Fei Liu. 2024{\natexlab{b}}.
\newblock \href {https://arxiv.org/abs/2406.12084} {When reasoning meets information aggregation: A case study with sports narratives}.
\newblock \emph{Preprint}, arXiv:2406.12084.

\bibitem[{Huang and Chang(2023)}]{huang2023reasoninglargelanguagemodels}
Jie Huang and Kevin Chen-Chuan Chang. 2023.
\newblock \href {https://arxiv.org/abs/2212.10403} {Towards reasoning in large language models: A survey}.
\newblock \emph{Preprint}, arXiv:2212.10403.

\bibitem[{Kavukcuoglu(2025)}]{google2025gemini25}
Koray Kavukcuoglu. 2025.
\newblock \href {https://blog.google/technology/google-deepmind/gemini-model-thinking-updates-march-2025/} {Gemini 2.5: Our most intelligent ai model}.
\newblock Technical report.
\newblock Accessed: 2025-05-13.

\bibitem[{Keith and Stent(2019)}]{keith-stent-2019-modeling}
Katherine Keith and Amanda Stent. 2019.
\newblock \href {https://doi.org/10.18653/v1/P19-1047} {Modeling financial analysts{'} decision making via the pragmatics and semantics of earnings calls}.
\newblock In \emph{Proceedings of the 57th Annual Meeting of the Association for Computational Linguistics}, pages 493--503, Florence, Italy. Association for Computational Linguistics.

\bibitem[{Khatuya et~al.(2024)Khatuya, Sinha, Ganguly, Ghosh, and Goyal}]{khatuya2024instructionguidedbulletpointsummarization}
Subhendu Khatuya, Koushiki Sinha, Niloy Ganguly, Saptarshi Ghosh, and Pawan Goyal. 2024.
\newblock \href {https://arxiv.org/abs/2405.06669} {Instruction-guided bullet point summarization of long financial earnings call transcripts}.
\newblock \emph{Preprint}, arXiv:2405.06669.

\bibitem[{Koa et~al.(2024)Koa, Ma, Ng, and Chua}]{Koa_2024}
Kelvin~J.L. Koa, Yunshan Ma, Ritchie Ng, and Tat-Seng Chua. 2024.
\newblock \href {https://doi.org/10.1145/3589334.3645611} {Learning to generate explainable stock predictions using self-reflective large language models}.
\newblock In \emph{Proceedings of the ACM Web Conference 2024}, volume 12706 of \emph{WWW ’24}, page 4304–4315. ACM.

\bibitem[{Krishna et~al.(2023)Krishna, Bransom, Kuehl, Iyyer, Dasigi, Cohan, and Lo}]{krishna-etal-2023-longeval}
Kalpesh Krishna, Erin Bransom, Bailey Kuehl, Mohit Iyyer, Pradeep Dasigi, Arman Cohan, and Kyle Lo. 2023.
\newblock \href {https://doi.org/10.18653/v1/2023.eacl-main.121} {{L}ong{E}val: Guidelines for human evaluation of faithfulness in long-form summarization}.
\newblock In \emph{Proceedings of the 17th Conference of the European Chapter of the Association for Computational Linguistics}, pages 1650--1669, Dubrovnik, Croatia. Association for Computational Linguistics.

\bibitem[{Laban et~al.(2024)Laban, Fabbri, Xiong, and Wu}]{laban2024summaryhaystackchallengelongcontext}
Philippe Laban, Alexander~R. Fabbri, Caiming Xiong, and Chien-Sheng Wu. 2024.
\newblock \href {https://arxiv.org/abs/2407.01370} {Summary of a haystack: A challenge to long-context llms and rag systems}.
\newblock \emph{Preprint}, arXiv:2407.01370.

\bibitem[{Lehman et~al.(2022)Lehman, Lialin, Legaspi, Sy, Pile, Alberto, Ragasa, Puyat, Tali{\~n}o, Alberto, Alfonso, Moukheiber, Wallace, Rumshisky, Liang, Raghavan, Celi, and Szolovits}]{lehman-etal-2022-learning}
Eric Lehman, Vladislav Lialin, Katelyn~Edelwina Legaspi, Anne~Janelle Sy, Patricia~Therese Pile, Nicole~Rose Alberto, Richard~Raymund Ragasa, Corinna~Victoria Puyat, Marianne~Katharina Tali{\~n}o, Isabelle~Rose Alberto, Pia~Gabrielle Alfonso, Dana Moukheiber, Byron Wallace, Anna Rumshisky, Jennifer Liang, Preethi Raghavan, Leo~Anthony Celi, and Peter Szolovits. 2022.
\newblock \href {https://doi.org/10.18653/v1/2022.clinicalnlp-1.8} {Learning to ask like a physician}.
\newblock In \emph{Proceedings of the 4th Clinical Natural Language Processing Workshop}, pages 74--86, Seattle, WA. Association for Computational Linguistics.

\bibitem[{Lewis and Mitchell(2024)}]{lewis2024usingcounterfactualtasksevaluate}
Martha Lewis and Melanie Mitchell. 2024.
\newblock \href {https://arxiv.org/abs/2402.08955} {Using counterfactual tasks to evaluate the generality of analogical reasoning in large language models}.
\newblock \emph{Preprint}, arXiv:2402.08955.

\bibitem[{Li et~al.(2025)Li, Cao, Griggs, Liu, Mo, Tang, Hegde, Hakhamaneshi, Patil, Zaharia, Gonzalez, and Stoica}]{li2025llmseasilylearnreason}
Dacheng Li, Shiyi Cao, Tyler Griggs, Shu Liu, Xiangxi Mo, Eric Tang, Sumanth Hegde, Kourosh Hakhamaneshi, Shishir~G. Patil, Matei Zaharia, Joseph~E. Gonzalez, and Ion Stoica. 2025.
\newblock \href {https://arxiv.org/abs/2502.07374} {Llms can easily learn to reason from demonstrations structure, not content, is what matters!}
\newblock \emph{Preprint}, arXiv:2502.07374.

\bibitem[{Li et~al.(2020)Li, Yang, Smyth, and Dong}]{CIKM2020MAEC}
Jiazheng Li, Linyi Yang, Barry Smyth, and Ruihai Dong. 2020.
\newblock \href {https://doi.org/10.1145/3340531.3412879} {Maec: A multimodal aligned earnings conference call dataset for financial risk prediction}.
\newblock In \emph{Proceedings of the 29th ACM International Conference on Information \&amp; Knowledge Management}, CIKM '20, page 3063–3070, New York, NY, USA. Association for Computing Machinery.

\bibitem[{Lin et~al.(2024)Lin, Trivedi, and Sun}]{lin2024generatingconfidenceuncertaintyquantification}
Zhen Lin, Shubhendu Trivedi, and Jimeng Sun. 2024.
\newblock \href {https://arxiv.org/abs/2305.19187} {Generating with confidence: Uncertainty quantification for black-box large language models}.
\newblock \emph{Preprint}, arXiv:2305.19187.

\bibitem[{Liu et~al.(2023)Liu, Lin, Hewitt, Paranjape, Bevilacqua, Petroni, and Liang}]{liu2023lostmiddlelanguagemodels}
Nelson~F. Liu, Kevin Lin, John Hewitt, Ashwin Paranjape, Michele Bevilacqua, Fabio Petroni, and Percy Liang. 2023.
\newblock \href {https://arxiv.org/abs/2307.03172} {Lost in the middle: How language models use long contexts}.
\newblock \emph{Preprint}, arXiv:2307.03172.

\bibitem[{Liu et~al.(2024)Liu, Fu, Yogatama, and Neiswanger}]{liu2024dellmaframeworkdecisionmaking}
Ollie Liu, Deqing Fu, Dani Yogatama, and Willie Neiswanger. 2024.
\newblock \href {https://arxiv.org/abs/2402.02392} {Dellma: A framework for decision making under uncertainty with large language models}.
\newblock \emph{Preprint}, arXiv:2402.02392.

\bibitem[{Loukas et~al.(2022)Loukas, Fergadiotis, Chalkidis, Spyropoulou, Malakasiotis, Androutsopoulos, and Paliouras}]{Loukas_2022}
Lefteris Loukas, Manos Fergadiotis, Ilias Chalkidis, Eirini Spyropoulou, Prodromos Malakasiotis, Ion Androutsopoulos, and Georgios Paliouras. 2022.
\newblock \href {https://doi.org/10.18653/v1/2022.acl-long.303} {Finer: Financial numeric entity recognition for xbrl tagging}.
\newblock In \emph{Proceedings of the 60th Annual Meeting of the Association for Computational Linguistics (Volume 1: Long Papers)}. Association for Computational Linguistics.

\bibitem[{Lu et~al.(2025)Lu, Hu, Foroosh, Jin, and Liu}]{lu-etal-2025-strux}
Yiming Lu, Yebowen Hu, Hassan Foroosh, Wei Jin, and Fei Liu. 2025.
\newblock \href {https://aclanthology.org/2025.naacl-short.11/} {{STRUX}: An {LLM} for decision-making with structured explanations}.
\newblock In \emph{Proceedings of the 2025 Conference of the Nations of the Americas Chapter of the Association for Computational Linguistics: Human Language Technologies (Volume 2: Short Papers)}, pages 131--141, Albuquerque, New Mexico. Association for Computational Linguistics.

\bibitem[{Medya et~al.(2022)Medya, Rasoolinejad, Yang, and Uzzi}]{medya2022exploratorystudystockprice}
Sourav Medya, Mohammad Rasoolinejad, Yang Yang, and Brian Uzzi. 2022.
\newblock \href {https://arxiv.org/abs/2203.12460} {An exploratory study of stock price movements from earnings calls}.
\newblock \emph{Preprint}, arXiv:2203.12460.

\bibitem[{Mondorf and Plank(2024)}]{mondorf2024accuracyevaluatingreasoningbehavior}
Philipp Mondorf and Barbara Plank. 2024.
\newblock \href {https://arxiv.org/abs/2404.01869} {Beyond accuracy: Evaluating the reasoning behavior of large language models -- a survey}.
\newblock \emph{Preprint}, arXiv:2404.01869.

\bibitem[{Mukherjee et~al.(2022)Mukherjee, Bohra, Banerjee, Sharma, Hegde, Shaikh, Shrivastava, Dasgupta, Ganguly, Ghosh, and Goyal}]{mukherjee2022ectsumnewbenchmarkdataset}
Rajdeep Mukherjee, Abhinav Bohra, Akash Banerjee, Soumya Sharma, Manjunath Hegde, Afreen Shaikh, Shivani Shrivastava, Koustuv Dasgupta, Niloy Ganguly, Saptarshi Ghosh, and Pawan Goyal. 2022.
\newblock \href {https://arxiv.org/abs/2210.12467} {Ectsum: A new benchmark dataset for bullet point summarization of long earnings call transcripts}.
\newblock \emph{Preprint}, arXiv:2210.12467.

\bibitem[{Musker et~al.(2024)Musker, Duchnowski, Millière, and Pavlick}]{musker2024semanticstructuremappingllmhuman}
Sam Musker, Alex Duchnowski, Raphaël Millière, and Ellie Pavlick. 2024.
\newblock \href {https://arxiv.org/abs/2406.13803} {Semantic structure-mapping in llm and human analogical reasoning}.
\newblock \emph{Preprint}, arXiv:2406.13803.

\bibitem[{Ni et~al.(2024)Ni, Meng, Chen, Zhao, Chen, Li, Zhang, Yin, Wang, and Chan}]{Ni_2024}
Haowei Ni, Shuchen Meng, Xupeng Chen, Ziqing Zhao, Andi Chen, Panfeng Li, Shiyao Zhang, Qifu Yin, Yuanqing Wang, and Yuxi Chan. 2024.
\newblock \href {https://doi.org/10.1109/docs63458.2024.10704454} {Harnessing earnings reports for stock predictions: A qlora-enhanced llm approach}.
\newblock In \emph{2024 6th International Conference on Data-driven Optimization of Complex Systems (DOCS)}, page 909–915. IEEE.

\bibitem[{Nie et~al.(2024)Nie, Kong, Dong, Mulvey, Poor, Wen, and Zohren}]{nie2024surveylargelanguagemodels}
Yuqi Nie, Yaxuan Kong, Xiaowen Dong, John~M. Mulvey, H.~Vincent Poor, Qingsong Wen, and Stefan Zohren. 2024.
\newblock \href {https://arxiv.org/abs/2406.11903} {A survey of large language models for financial applications: Progress, prospects and challenges}.
\newblock \emph{Preprint}, arXiv:2406.11903.

\bibitem[{OpenAI et~al.(2024)OpenAI, :, Jaech, Kalai, Lerer, Richardson, El-Kishky, Low, Helyar, Madry, Beutel, Carney, Iftimie, Karpenko, Passos, Neitz, Prokofiev, Wei, Tam, Bennett, Kumar, Saraiva, Vallone, Duberstein, Kondrich, Mishchenko, Applebaum, Jiang, Nair, Zoph, Ghorbani, Rossen, Sokolowsky, Barak, McGrew, Minaiev, Hao, Baker, Houghton, McKinzie, Eastman, Lugaresi, Bassin, Hudson, Li, de~Bourcy, Voss, Shen, Zhang, Koch, Orsinger, Hesse, Fischer, Chan, Roberts, Kappler, Levy, Selsam, Dohan, Farhi, Mely, Robinson, Tsipras, Li, Oprica, Freeman, Zhang, Wong, Proehl, Cheung, Mitchell, Wallace, Ritter, Mays, Wang, Such, Raso, Leoni, Tsimpourlas, Song, von Lohmann, Sulit, Salmon, Parascandolo, Chabot, Zhao, Brockman, Leclerc, Salman, Bao, Sheng, Andrin, Bagherinezhad, Ren, Lightman, Chung, Kivlichan, O'Connell, Osband, Gilaberte, Akkaya, Kostrikov, Sutskever, Kofman, Pachocki, Lennon, Wei, Harb, Twore, Feng, Yu, Weng, Tang, Yu, Candela, Palermo, Parish, Heidecke, Hallman, Rizzo, Gordon, Uesato, Ward, Huizinga, Wang, Chen, Xiao, Singhal, Nguyen, Cobbe, Shi, Wood, Rimbach, Gu-Lemberg, Liu, Lu, Stone, Yu, Ahmad, Yang, Liu, Maksin, Ho, Fedus, Weng, Li, McCallum, Held, Kuhn, Kondraciuk, Kaiser, Metz, Boyd, Trebacz, Joglekar, Chen, Tintor, Meyer, Jones, Kaufer, Schwarzer, Shah, Yatbaz, Guan, Xu, Yan, Glaese, Chen, Lampe, Malek, Wang, Fradin, McClay, Pavlov, Wang, Wang, Murati, Bavarian, Rohaninejad, McAleese, Chowdhury, Chowdhury, Ryder, Tezak, Brown, Nachum, Boiko, Murk, Watkins, Chao, Ashbourne, Izmailov, Zhokhov, Dias, Arora, Lin, Lopes, Gaon, Miyara, Leike, Hwang, Garg, Brown, James, Shu, Cheu, Greene, Jain, Altman, Toizer, Toyer, Miserendino, Agarwal, Hernandez, Baker, McKinney, Yan, Zhao, Hu, Santurkar, Chaudhuri, Zhang, Fu, Papay, Lin, Balaji, Sanjeev, Sidor, Broda, Clark, Wang, Gordon, Sanders, Patwardhan, Sottiaux, Degry, Dimson, Zheng, Garipov, Stasi, Bansal, Creech, Peterson, Eloundou, Qi, Kosaraju, Monaco, Pong, Fomenko, Zheng, Zhou, McCabe, Zaremba, Dubois, Lu, Chen, Cha, Bai, He, Zhang, Wang, Shao, and Li}]{openai2024openaio1card}
OpenAI, :, Aaron Jaech, Adam Kalai, Adam Lerer, Adam Richardson, Ahmed El-Kishky, Aiden Low, Alec Helyar, Aleksander Madry, Alex Beutel, Alex Carney, Alex Iftimie, Alex Karpenko, Alex~Tachard Passos, Alexander Neitz, Alexander Prokofiev, Alexander Wei, Allison Tam, and 244 others. 2024.
\newblock \href {https://arxiv.org/abs/2412.16720} {Openai o1 system card}.
\newblock \emph{Preprint}, arXiv:2412.16720.

\bibitem[{Ozturkler et~al.(2023)Ozturkler, Malkin, Wang, and Jojic}]{ozturkler2023thinksumprobabilisticreasoningsets}
Batu Ozturkler, Nikolay Malkin, Zhen Wang, and Nebojsa Jojic. 2023.
\newblock \href {https://arxiv.org/abs/2210.01293} {Thinksum: Probabilistic reasoning over sets using large language models}.
\newblock \emph{Preprint}, arXiv:2210.01293.

\bibitem[{Peters(2024)}]{hsgac2024aihedgefunds}
Gary Peters. 2024.
\newblock \href {https://www.hsgac.senate.gov/wp-content/uploads/2024.06.11-Hedge-Fund-Use-of-AI-Report.pdf} {Hedge funds' use of artificial intelligence and machine learning technologies}.
\newblock Technical report, U.S. Senate Committee on Homeland Security and Governmental Affairs.
\newblock Accessed: 2025-05-12.

\bibitem[{Qin et~al.(2024)Qin, Xia, Wang, Jiao, Hu, Ding, Chen, and Joty}]{qin2024relevantrandomllmstruly}
Chengwei Qin, Wenhan Xia, Tan Wang, Fangkai Jiao, Yuchen Hu, Bosheng Ding, Ruirui Chen, and Shafiq Joty. 2024.
\newblock \href {https://arxiv.org/abs/2404.12728} {Relevant or random: Can llms truly perform analogical reasoning?}
\newblock \emph{Preprint}, arXiv:2404.12728.

\bibitem[{Reddy et~al.(2024)Reddy, Koncel-Kedziorski, Lai, Krumdick, Lovering, and Tanner}]{reddy2024docfinqa}
Varshini Reddy, Rik Koncel-Kedziorski, Viet~Dac Lai, Michael Krumdick, Charles Lovering, and Chris Tanner. 2024.
\newblock \href {https://arxiv.org/abs/2401.06915} {Docfinqa: A long-context financial reasoning dataset}.
\newblock \emph{Preprint}, arXiv:2401.06915.

\bibitem[{Ren et~al.(2025)Ren, Shao, Song, Xin, Wang, Zhao, Zhang, Fu, Zhu, Yang, Wu, Gou, Ma, Tang, Liu, Gao, Guo, and Ruan}]{ren2025deepseekproverv2advancingformalmathematical}
Z.~Z. Ren, Zhihong Shao, Junxiao Song, Huajian Xin, Haocheng Wang, Wanjia Zhao, Liyue Zhang, Zhe Fu, Qihao Zhu, Dejian Yang, Z.~F. Wu, Zhibin Gou, Shirong Ma, Hongxuan Tang, Yuxuan Liu, Wenjun Gao, Daya Guo, and Chong Ruan. 2025.
\newblock \href {https://arxiv.org/abs/2504.21801} {Deepseek-prover-v2: Advancing formal mathematical reasoning via reinforcement learning for subgoal decomposition}.
\newblock \emph{Preprint}, arXiv:2504.21801.

\bibitem[{Sang and Bao(2022)}]{sang-bao-2022-dialoguegat}
Yunxin Sang and Yang Bao. 2022.
\newblock \href {https://doi.org/10.18653/v1/2022.findings-emnlp.117} {{D}ialogue{GAT}: A graph attention network for financial risk prediction by modeling the dialogues in earnings conference calls}.
\newblock In \emph{Findings of the Association for Computational Linguistics: EMNLP 2022}, pages 1623--1633, Abu Dhabi, United Arab Emirates. Association for Computational Linguistics.

\bibitem[{Sawhney et~al.(2021)Sawhney, Goyal, Goel, Mathur, and Shah}]{sawhney-etal-2021-multimodal}
Ramit Sawhney, Mihir Goyal, Prakhar Goel, Puneet Mathur, and Rajiv~Ratn Shah. 2021.
\newblock \href {https://doi.org/10.18653/v1/2021.acl-long.526} {Multimodal multi-speaker merger {\&} acquisition financial modeling: A new task, dataset, and neural baselines}.
\newblock In \emph{Proceedings of the 59th Annual Meeting of the Association for Computational Linguistics and the 11th International Joint Conference on Natural Language Processing (Volume 1: Long Papers)}, pages 6751--6762, Online. Association for Computational Linguistics.

\bibitem[{Sawhney et~al.(2020)Sawhney, Khanna, Aggarwal, Jain, Mathur, and Shah}]{sawhney-etal-2020-voltage}
Ramit Sawhney, Piyush Khanna, Arshiya Aggarwal, Taru Jain, Puneet Mathur, and Rajiv~Ratn Shah. 2020.
\newblock \href {https://doi.org/10.18653/v1/2020.emnlp-main.643} {{V}ol{TAGE}: Volatility forecasting via text audio fusion with graph convolution networks for earnings calls}.
\newblock In \emph{Proceedings of the 2020 Conference on Empirical Methods in Natural Language Processing (EMNLP)}, pages 8001--8013, Online. Association for Computational Linguistics.

\bibitem[{Sonkiya et~al.(2021)Sonkiya, Bajpai, and Bansal}]{sonkiya2021stockpricepredictionusing}
Priyank Sonkiya, Vikas Bajpai, and Anukriti Bansal. 2021.
\newblock \href {https://arxiv.org/abs/2107.09055} {Stock price prediction using bert and gan}.
\newblock \emph{Preprint}, arXiv:2107.09055.

\bibitem[{Sourati et~al.(2024)Sourati, Ilievski, Sommerauer, and Jiang}]{sourati2024arnanalogicalreasoningnarratives}
Zhivar Sourati, Filip Ilievski, Pia Sommerauer, and Yifan Jiang. 2024.
\newblock \href {https://arxiv.org/abs/2310.00996} {Arn: Analogical reasoning on narratives}.
\newblock \emph{Preprint}, arXiv:2310.00996.

\bibitem[{Sprague et~al.(2024)Sprague, Ye, Bostrom, Chaudhuri, and Durrett}]{sprague2024musr}
Zayne Sprague, Xi~Ye, Kaj Bostrom, Swarat Chaudhuri, and Greg Durrett. 2024.
\newblock \href {https://arxiv.org/abs/2310.16049} {Musr: Testing the limits of chain-of-thought with multistep soft reasoning}.
\newblock \emph{Preprint}, arXiv:2310.16049.

\bibitem[{Srivastava et~al.(2024)Srivastava, Malik, Gupta, Ganu, and Roth}]{srivastava2024evaluatingllmsmathematicalreasoning}
Pragya Srivastava, Manuj Malik, Vivek Gupta, Tanuja Ganu, and Dan Roth. 2024.
\newblock \href {https://arxiv.org/abs/2402.11194} {Evaluating llms' mathematical reasoning in financial document question answering}.
\newblock \emph{Preprint}, arXiv:2402.11194.

\bibitem[{Wang et~al.(2023)Wang, Li, Zhao, Kou, Wang, Zhu, Wang, Shen, and Chen}]{wang2023methodsacquiringincorporatingknowledge}
Liping Wang, Jiawei Li, Lifan Zhao, Zhizhuo Kou, Xiaohan Wang, Xinyi Zhu, Hao Wang, Yanyan Shen, and Lei Chen. 2023.
\newblock \href {https://arxiv.org/abs/2308.04947} {Methods for acquiring and incorporating knowledge into stock price prediction: A survey}.
\newblock \emph{Preprint}, arXiv:2308.04947.

\bibitem[{Webb et~al.(2023)Webb, Holyoak, and Lu}]{webb2023emergentanalogicalreasoninglarge}
Taylor Webb, Keith~J. Holyoak, and Hongjing Lu. 2023.
\newblock \href {https://arxiv.org/abs/2212.09196} {Emergent analogical reasoning in large language models}.
\newblock \emph{Preprint}, arXiv:2212.09196.

\bibitem[{Yang et~al.(2024)Yang, Dailisan, Korecki, Hausladen, and Helbing}]{yang2024llmvotinghumanchoices}
Joshua~C. Yang, Damian Dailisan, Marcin Korecki, Carina~I. Hausladen, and Dirk Helbing. 2024.
\newblock \href {https://arxiv.org/abs/2402.01766} {Llm voting: Human choices and ai collective decision making}.
\newblock \emph{Preprint}, arXiv:2402.01766.

\bibitem[{Yasunaga et~al.(2024)Yasunaga, Chen, Li, Pasupat, Leskovec, Liang, Chi, and Zhou}]{yasunaga2024largelanguagemodelsanalogical}
Michihiro Yasunaga, Xinyun Chen, Yujia Li, Panupong Pasupat, Jure Leskovec, Percy Liang, Ed~H. Chi, and Denny Zhou. 2024.
\newblock \href {https://arxiv.org/abs/2310.01714} {Large language models as analogical reasoners}.
\newblock \emph{Preprint}, arXiv:2310.01714.

\bibitem[{Ye et~al.(2024)Ye, Cong, Tian, Qin, Liu, Lin, Liu, and Sun}]{ye2024rationaldecisionmakingagentinternalized}
Yining Ye, Xin Cong, Shizuo Tian, Yujia Qin, Chong Liu, Yankai Lin, Zhiyuan Liu, and Maosong Sun. 2024.
\newblock \href {https://arxiv.org/abs/2308.12519} {Rational decision-making agent with internalized utility judgment}.
\newblock \emph{Preprint}, arXiv:2308.12519.

\bibitem[{Yu et~al.(2024)Yu, He, and Ying}]{yu2024thoughtpropagationanalogicalapproach}
Junchi Yu, Ran He, and Rex Ying. 2024.
\newblock \href {https://arxiv.org/abs/2310.03965} {Thought propagation: An analogical approach to complex reasoning with large language models}.
\newblock \emph{Preprint}, arXiv:2310.03965.

\bibitem[{Yuan et~al.(2024)Yuan, Chen, Sun, Liang, Xiao, and Yang}]{yuan2024analogykbunlockinganalogicalreasoning}
Siyu Yuan, Jiangjie Chen, Changzhi Sun, Jiaqing Liang, Yanghua Xiao, and Deqing Yang. 2024.
\newblock \href {https://arxiv.org/abs/2305.05994} {Analogykb: Unlocking analogical reasoning of language models with a million-scale knowledge base}.
\newblock \emph{Preprint}, arXiv:2305.05994.

\bibitem[{Zhu et~al.(2024)Zhu, Jiao, and Jordan}]{zhu2024principledreinforcementlearninghuman}
Banghua Zhu, Jiantao Jiao, and Michael~I. Jordan. 2024.
\newblock \href {https://arxiv.org/abs/2301.11270} {Principled reinforcement learning with human feedback from pairwise or $k$-wise comparisons}.
\newblock \emph{Preprint}, arXiv:2301.11270.

\bibitem[{Zhu et~al.(2021)Zhu, Lei, Huang, Wang, Zhang, Lv, Feng, and Chua}]{zhu2021tatqa}
Fengbin Zhu, Wenqiang Lei, Youcheng Huang, Chao Wang, Shuo Zhang, Jiancheng Lv, Fuli Feng, and Tat-Seng Chua. 2021.
\newblock \href {https://arxiv.org/abs/2105.07624} {Tat-qa: A question answering benchmark on a hybrid of tabular and textual content in finance}.
\newblock \emph{Preprint}, arXiv:2105.07624.

\end{thebibliography}
\appendix

\section{Influential Factors}
\label{sec:appendix}

\begin{itemize}[topsep=3pt,itemsep=5pt]
  
\item \textbf{Macroeconomic Influences.} These encompass broad economic factors that affect the entire market or large segments of it. This includes the overall economic health, market sentiment, political events, natural disasters and geopolitical issues~\citep{liu2024dellmaframeworkdecisionmaking}. Each factor leads to two potential outcomes; for instance, natural disasters might cause a `Major Impact' by disrupting economies and global supply chains, and directly affecting market performance; the `Unknown or Uncertain' outcome reflects the unpredictability of such events.

\item \textbf{Company-Specific Dynamics.} These factors are linked to the internal operations and strategic decisions of individual companies, such as mergers and acquisitions, regulatory changes, financial health, company growth potential, product launches, and issues within the supply chain. Each factor can result in one of two potential outcomes. For example, a `Positive Outlook' on regulatory changes can open up new business opportunities, whereas `Unknown or Uncertain' could signify regulatory uncertainties that lead to financial challenges.

\item \textbf{Historical Financial Metrics.} Important metrics include historical earnings per share (EPS), revenue trends, and past stock price movements. Each factor can result in three outcomes: `Bullish', where metrics like earnings per share, revenue, and stock prices consistently rise, indicating strong financial health; `Stable', characterized by steady movements; `Bearish,' marked by declining financial figures, possibly leading investors to be pessimistic about the company's future performance. 

\end{itemize}

\begin{figure*}[htbp]
\centering
\begin{footnotesize}
\begin{minipage}{\textwidth}

\begin{promptbox}{\textsf{\textbf{Transcript Excerpt from Delta Air Lines (DAL) Earnings Call}}}
\begin{lstlisting}[basicstyle=\footnotesize\setstretch{0.92}]
[Prepared Remarks:]

>> Operator

Good morning, everyone, and welcome to the Delta Air Lines September-quarter 2021 financial results conference call. My name is Jen, and I will be your coordinator. [Operator instructions] As a reminder, today's call is being recorded. I would now like to turn the conference over to Ms. Julie Stewart, vice president of investor relations...

>> Julie Stewart -- Vice President of Investor Relations

Thank you, Jen. Good morning, everyone, and thanks for joining us for our September-quarter 2021 earnings call. Joining us from Atlanta today are CEO, Ed Bastian; our president, Glen Hauenstein; our CFO, Dan Janki. And Ed will open the call with an overview of Delta's performance and strategy.

Glen will provide an update on the revenue environment and our brand momentum, and Dan will discuss cost fleet and our balance sheet. Similar to last quarter's call, we've scheduled today's call for 90 minutes to make sure we have plenty of time for questions. [Operator instructions] After the analyst Q\&A, we will move to our media questions, after which, Ed will provide a brief closing statement. Today's discussion contains forward-looking statements that represent our beliefs or expectations about future events.

All forward-looking statements involve risks and uncertainties that could cause the actual results to differ materially from the forward-looking statements. Some of the factors that may cause such differences are described in Delta's SEC filings. We also discuss non-GAAP financial measures, and all results exclude special items unless otherwise noted. You can find a reconciliation of our non-GAAP measures on the Investor Relations page at ir.delta.com. And with that, I'll turn the call over to Ed.

>> Ed Bastian -- Chief Executive Officer

Well, thank you, Julie, and good morning, everyone. Appreciate you joining us this morning. The September quarter marked another important milestone in our recovery. We achieved our first quarterly profit since the start of the pandemic with a pre-tax result of $216 million and a pre-tax margin of nearly 3% despite still missing one-third of our revenue base compared to the same period in 2019... [omitted.]

[Questions & Answers:]

>> Operator

Thank you. And we'll go first to Jamie Baker with J.P. Morgan.

>> Jamie Baker -- J.P. Morgan -- Analyst

Hey. Good morning, everybody. First question goes potentially to Glen and Dan. So pre-COVID, I had asked Paul about the amount of time that it would typically take Delta to recalibrate the higher fuel prices.

I'm not staring at the transcript, but his estimate at the time was four to six months, which was an improvement from historic levels. So my question, I guess, for Glen is whether the booking curve is steep enough right now that you might actually be able to recapture the top line more quickly than that. And similarly, for Dan, whether there's anything we should be taking on the cost or operations side that could accelerate the process. I'm basically just trying to understand whether four to six months is still the right estimate for us to be using.

>> Glen Hauenstein -- President

Well, I would just comment, I think we're a bit in uncharted territory here as the recovery continues. And while I think it might be difficult in the very short run, despite the fact that the booking curve has moved in a bit, that I would estimate that, that four to six months is about right because we believe that demand and capacity will fall back into a very good equilibrium by next spring which would put you inside that window... 

\end{lstlisting}

\end{promptbox}

\end{minipage}
\end{footnotesize}
\vspace{-0.1in}
\caption{Transcript Excerpt from Delta Air Lines (DAL) Earnings Call.
Adapted from~\cite{lu-etal-2025-strux}.
}
\label{fig:excerpt}
\end{figure*}

\begin{figure*}[htbp]
\centering
\begin{footnotesize}
\begin{minipage}{\textwidth}

\begin{promptbox}{\textsf{\textbf{Building a Factor Profile from the Earnings Call Transcript}}}

\textbf{\underline{System Message}}\\[0.5em]
You are a financial analyst specializing in earnings call transcripts. You will receive the complete transcript of an earnings call, which includes both the prepared remarks and the Q\&A session. Your job is to identify the key factors from the transcript and assign probabilities to the potential outcomes of these factors.\\[1em]

\textbf{\underline{User Message}}\\[0.5em]
Your task is to conduct a comprehensive analysis of the earnings call transcript below. Be sure to accurately capture the important factors and estimate the likelihood of each factor resulting in specific outcomes.\\[1em]

    Earnings Call Transcript for Company \{Company\}\\[1em]
    \# Prepared Remarks\\
    Speaker: \{Speech...\}\\
    ...\\[0.5em]
    \# Questions and Answers\\
    Analyst: \{Speech...\}\\
    ...\\[1em]
    
    Please analyze the above earnings call transcript, focusing on the following key factors:\\
    \{Enumerate factors, descriptions, and outcomes\}\\[1em]
    1. Economic health: Economic health refers to the overall stability and performance of the economy, reflected in factors like growth, employment, inflation, and market confidence. Outcomes: \{positive-outlook, unknown-or-uncertain\}\\[1em]
    2. Market sentiment and investor psychology: Market sentiment reflects the overall mood or attitude of investors toward a particular market, influenced by news, economic data, and global events. Investor psychology refers to the emotions and cognitive biases that drive decisions, often leading to behaviors like fear-driven selling or greed-fueled buying. Outcomes: \{optimistic, unknown-or-uncertain\}\\
    ...\\[1em]
    Please take the time to thoroughly understand the transcript. For each key factor, provide a detailed summary based on the given transcript. Then, review all associated outcomes and assess the likelihood of each outcome. The likelihood should be strictly selected from the following options: \{very likely, likely, somewhat likely, somewhat unlikely, unlikely, very unlikely\}. Format your response in JSON.\\[1em]
    
    \# Example Output:\\
    \{JSON output example\}\\[0.5em]
    \# Your Output:\\

\end{promptbox}

\end{minipage}
\end{footnotesize}
\vspace{-0.1in}
\caption{Building a Factor Profile from the Earnings Call Transcript}
\label{fig:factor_profile_prompt}
\end{figure*}

\begin{figure*}[htbp]
\centering
\begin{footnotesize}
\begin{minipage}{\textwidth}

\begin{promptbox}{\textsf{\textbf{Applying Analogical Reasoning to Investment Decisions}}}

\textbf{\underline{System Message}}\\[0.5em]
You're a financial analyst who specializes in giving investors buy or sell recommendations by thoroughly analyzing earnings call transcripts.\\[1em]

\textbf{\underline{User Message}}\\[0.5em]
Here are several example company profiles. Each profile highlights key factors from an earnings call transcript and probabilities for potential outcomes based on those factors. Each profile represents a specific company and is based on its historical earnings call data. Your job is to pick the most analogous example and use its strategy to solve the initial problem.\\[1em]

  Example Company Profile 1:\\
  \{Factor Profile 1\}\\
  Analyst recommendation: \{Action 1\}\\[0.5em]
  Example Company Profile 2:\\
  \{Factor Profile 2\}\\
  Analyst recommendation: \{Action 2\}\\[0.5em]
  Example Company Profile 3:\\
  \{Factor Profile 3\}\\
  Analyst recommendation: \{Action 3\}\\[0.5em]
  Example Company Profile 4:\\
  \{Factor Profile 4\}\\
  Analyst recommendation: \{Action 4\}\\[0.5em]
  Example Company Profile 5:\\
  \{Factor Profile 5\}\\
  Analyst recommendation: \{Action 5\}\\[1em]
  
  \textbf{** Initial Problem **}\\[0.5em]
  Based on your analysis of the earnings call for \{Company Name\} held on \{Announcement Date\}, decide on the most likely analyst recommendation for the next 30 days from these options: \\[1em]
  
  \quad - Action 1: strong buy: The stock price will increase by more than 5\%\\
  \quad - Action 2: buy: The stock price will increase by 2\% to 5\%\\
  \quad - Action 3: hold: The stock price is expected to remain stable, fluctuating between -2\% to 2\%\\
  \quad - Action 4: sell: The stock price will decrease by 2\% to 5\%\\
  \quad - Action 5: strong sell: The stock price will decrease by more than 5\%\\[1em]
  
  Below is the company profile summarized from \{Company Name\}'s earnings call on \{Announcement Date\} and the historical price trend probabilities judged by an analyst:\\[1em]
  
  \{Factor Profile Constructed Using an Earnings Call Transcript\}\\[1em]
  
  \textbf{** Solve the Initial Problem **}\\[1em]
  
  Please respond with the analyst recommendation for this stock in JSON format, including these keys: (`idx', `recommendation', `justification'). `idx' is the index of the most analogous example profile, and `recommendation' should be one of the actions mentioned above for 30 days of trading, and `justification' should clearly explain your recommendation using the strategy you learned from the selected example company profile.\\[1em]

  \end{promptbox}

\end{minipage}
\end{footnotesize}
\vspace{-0.1in}
\caption{Applying Analogical Reasoning to Investment Decisions.}
\label{fig:ar_prompt}
\end{figure*}

\begin{figure*}[htbp]
\centering
\begin{footnotesize}
\begin{minipage}{\textwidth}

\begin{promptbox}{\textsf{\textbf{Chain-of-Thought Prompt for Investment Decisions}}}

\textbf{\underline{System Message}}\\[0.5em]
You're a financial analyst spcializing in giving investors buy or sell recommendations by thoroughly analyzing earnings call transcripts.\\[1em]

\textbf{\underline{User Message}}\\[0.5em]
Based on your analysis of the earnings call for \{Company Name\} held on \{Announcement Date\}, decide on the most likely analyst recommendation for the next 30 days from these options: \\[1em]

  \quad - Action 1: strong buy: The stock price will increase by more than 5\%\\
  \quad - Action 2: buy: The stock price will increase by 2\% to 5\%\\
  \quad - Action 3: hold: The stock price is expected to remain stable, fluctuating between -2\% to 2\%\\
  \quad - Action 4: sell: The stock price will decrease by 2\% to 5\%\\
  \quad - Action 5: strong sell: The stock price will decrease by more than 5\%\\[1em]
  
  Below is the \{Factor Profile, Transcript or Summary\} from \{Company Name\}'s earnings call on \{Announcement Date\}:\}\\[1em]
  \{Factor Profile, Transcripts or Summary\}\\[1em]
  
  Please think step by step and respond with the analyst recommendation for this stock in JSON format, including these keys: (`thoughts', `recommendation', `justification'). `Thoughts' should be your detailed reasoning steps, `recommendation' should be one of the actions mentioned above for 30 days trading, `Justification' should clearly explain your recommendation using the strategy you learned from the selected example company profile.\\[1em]

  \end{promptbox}

\end{minipage}
\end{footnotesize}
\vspace{-0.1in}
\caption{Chain-of-Thought Prompt for Soliciting Investment Decisions.}
\label{fig:cot_prompt}
\end{figure*}

\begin{figure*}[htbp]
\centering
\begin{footnotesize}
\begin{minipage}{\textwidth}

\begin{promptbox}{\textsf{\textbf{Prompt for Analyzing Trends in Historical Financial Data}}}

\textbf{\underline{System Message}}\\[0.5em]
You are a financial analyst specializing in historical data analysis, including stock prices, earnings per share (EPS), and revenue. Your goal is to assess the likelihood of different market trends based on past data.\\[1em]

\textbf{\underline{User Message}}\\[0.5em]
The potential outcomes to consider are: \{bullish, stable, and bearish\}. For each outcome, please assign a likelihood level from the following options: \{very likely, likely, somewhat likely, somewhat unlikely, unlikely, very unlikely\}.\\[1em]

    Below, you will be provided with a historical data table:
    \{Data Name\}:\{Description\}\\
    \{Historical Data Table\}\\[1em]
    
    \text{\quad Date \,\,\,\,\,\,\,\,\,\,\,\,\,\,\,\quad Close Price} \\
    \text{\quad 2023-07-31 \quad 195.22} \\
    \text{\quad 2023-08-01 \quad 195.46} \\
    \{...until the date of the earnings announcement.\}\\[1em]
    
    Please analyze this historical data and provide the likelihood of each outcome in JSON format. \\[1em]
    \# Example Output:\\
    \{"historical EPS":\{"bullish": very likely, "stable":somewhat likely, "bearish": unlikely\}\}\\[0.5em]
    \# Your Output:

  \end{promptbox}

\end{minipage}
\end{footnotesize}
\vspace{-0.1in}
\caption{Prompt for Analyzing Trends in Historical Financial Data.}
\label{fig:historical_analyze_prompt}
\end{figure*}

\end{document}